\documentclass{article}

\usepackage[final,main]{neurips_2026}

\usepackage[utf8]{inputenc}
\usepackage[T1]{fontenc}
\usepackage{hyperref}
\usepackage{url}
\usepackage{booktabs}
\usepackage{amsfonts}
\usepackage{amsmath}
\usepackage{amssymb}
\usepackage{nicefrac}
\usepackage{microtype}
\usepackage{xcolor}
\usepackage{graphicx}
\usepackage{multirow}
\usepackage{makecell}
\usepackage{caption}
\usepackage{wrapfig}
\usepackage{placeins}

\hypersetup{colorlinks=true,linkcolor=black,citecolor=black,urlcolor=black}

\newcommand{\R}{\mathbb{R}}
\newcommand{\cA}{\mathcal{A}}
\newcommand{\htil}{\tilde{h}}
\newcommand{\arrhat}{\widehat{\mathrm{arr}}}
\newcommand{\dephat}{\widehat{\mathrm{dep}}}
\newcommand{\arr}{\mathrm{arr}}
\newcommand{\dep}{\mathrm{dep}}

\title{LINC: Decoupling Local Consequence Scoring from\\ Hidden Matching in Constructive Neural Routing}

\author{%
  \textbf{ShaoFeng Qin, Li Wang}\thanks{Corresponding author. Code: \url{https://github.com/Elaina10172004/LINC}}\\
  \textbf{Beijing University of Posts and Telecommunications}\\
  \texttt{elaina20041017@gmail.com} \quad \texttt{wang\_li@bupt.edu.cn}
}

\begin{document}

\maketitle

\begin{abstract}
Constructive neural routing solvers usually score the next action by matching a decoder context against candidate embeddings computed once per instance from static node features such as coordinates, demands, and time windows; deterministic one-step consequences such as travel, waiting, slack, and capacity changes are not provided as inputs and must be inferred implicitly by the model. We propose LINC, a decoder-side candidate decision architecture that decouples computing these consequences from learning how much they matter. The environment computes the consequences explicitly at each step; candidate-level consequences are scored by a state-conditioned shared linear comparator, while feasible-set summaries modulate the decoder context. This keeps the cached embeddings and standard global matching while giving the scoring function direct access to deterministic transition quantities. The Capacitated Vehicle Routing Problem with Time Windows (CVRPTW) serves as the main constrained-routing testbed, and the same interface extends to the Capacitated Vehicle Routing Problem (CVRP) and Traveling Salesman Problem (TSP). Across external benchmarks and no-retraining scale-transfer settings, LINC consistently improves strong neural baselines, with the advantage becoming more pronounced as test size moves further beyond the training scale, especially on constrained routing problems.
\end{abstract}

\section{Introduction}

Constructive neural combinatorial optimization (NCO) has become a standard learning-based paradigm for routing problems: a policy builds a solution one action at a time, and a decoder scores the feasible next nodes conditioned on the partial tour. Starting from pointer-network and attention-model formulations \citep{vinyals2015pointer,bello2017neural,nazari2018reinforcement,kool2019attention}, later work improved exploration and inference through multi-start rollouts, population policies, and guided search \citep{kwon2020pomo,grinsztajn2023winner,chalumeau2023combinatorial,hottung2025polynet,choo2022simulation}. These techniques have made neural constructors competitive on many synthetic TSP and CVRP settings.

Constructive routing requires comparing candidates with \emph{state-dependent consequences}. In CVRPTW, picking a customer determines arrival time, waiting time, service start, departure, remaining capacity, and the updated time budget constraining later feasibility checks. These quantities follow deterministically from the MDP transition. Although the environment uses them to enforce feasibility, conventional attention-based scoring does not directly expose them as candidate-level inputs. The policy instead ranks feasible actions through learned interactions between a decoder context and cached node representations. We hypothesize that providing these consequences explicitly can reduce the representational burden of learning state-dependent trade-offs, particularly under distribution shift.

\begin{wrapfigure}{r}{0.4\textwidth}
  \vspace{-12pt}
  \centering
  \includegraphics[width=0.39\textwidth]{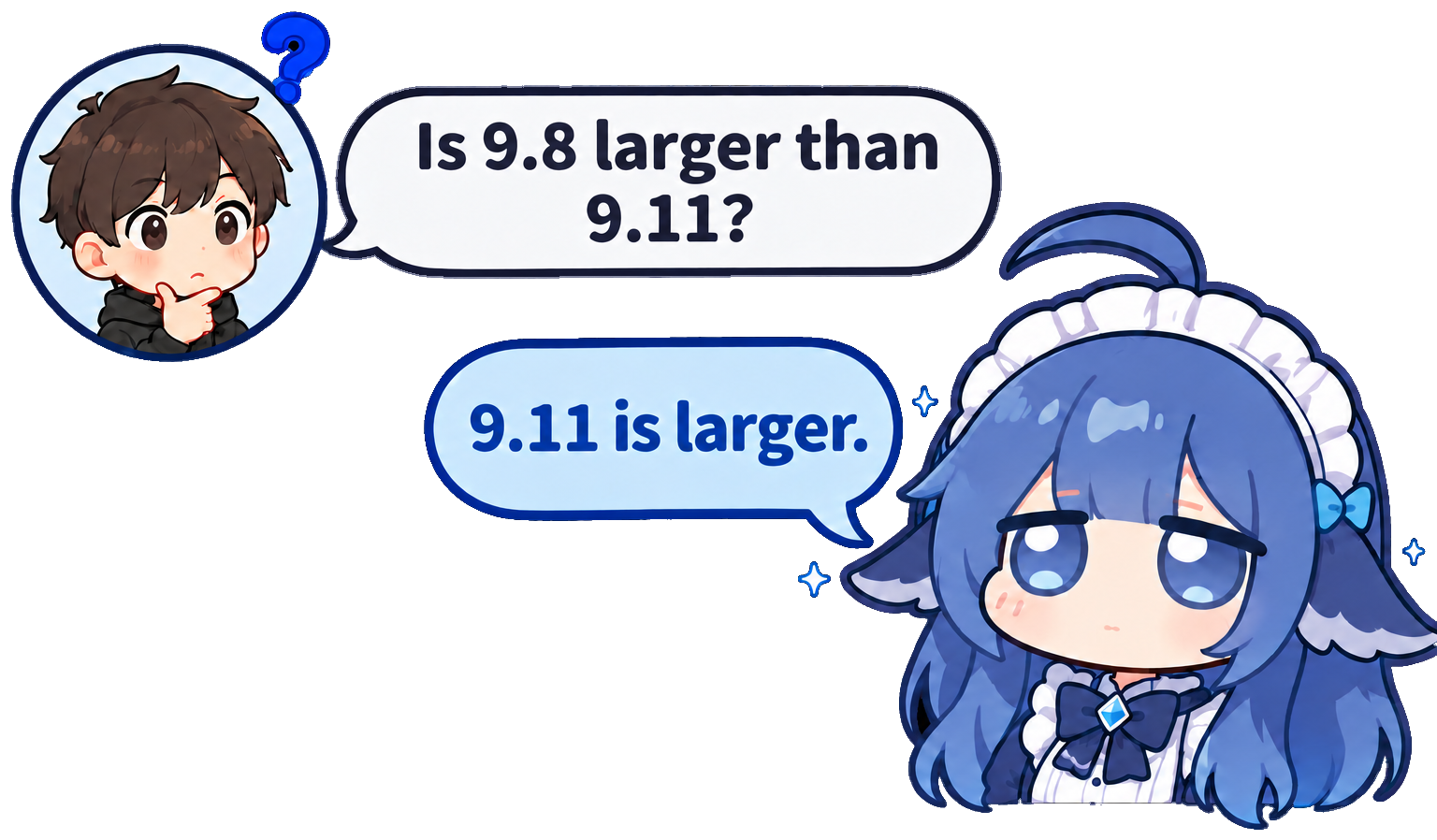}
  \caption{Even very large language models can fail at a one-step exact comparison; such computations are usually delegated to an external program.}
  \label{fig:llm}
  \vspace{-10pt}
\end{wrapfigure}
A related design choice appears in language-model systems: deterministic computations can be delegated to an external program whose result is placed in the context \citep{schick2023toolformer,gao2023pal}, avoiding errors such as the comparison illustrated in Figure~\ref{fig:llm}. This motivates giving routing policies direct access to quantities that can be computed exactly at low cost.

This paper argues that candidate scoring should separate \emph{computing local consequences} from \emph{learning their decision relevance}. The environment already determines how a candidate changes travel, waiting, timing slack, and capacity. The policy must learn how to weigh these quantities against global routing strategy. In the Attention Model (AM) \citep{kool2019attention}, POMO, and PolyNet, the encoder typically runs once per instance to produce a representation of each node, which is cached and shared by all rollouts; at each step the decoder only updates its query and scores candidates by their inner products with these node representations. Because the consequences change with the current state, incorporating them through the encoder would require updating representations during construction, sacrificing the reuse of a single static encoding. We instead retain the cached candidate keys and score the explicit consequences through a shared linear comparator. Its coefficients depend on the decoder state and feasible-set summaries, allowing the same candidate quantities to receive different weights as the routing state changes. This comparison-and-modulation structure defines how the dynamic information enters the decision.

We propose LINC, a candidate decision architecture for constructive routing. LINC explicitly computes local consequence features for each feasible candidate, scores them with a state-conditioned shared comparator, summarizes the step-level candidate set, and combines the result with standard context-key attention. The resulting decoder combines global matching through learned embeddings with direct, state-dependent comparison of explicit transition quantities.

CVRPTW is the primary testbed because it couples geometry, capacity, and temporal feasibility; TSP and CVRP are included to test whether the same interface remains useful when constraints are simpler. The strongest empirical gains occur on CVRPTW: on Solomon and Homberger external benchmarks, LINC reduces the unguided gap from 13.8\%/38.14\% to 7.3\%/14.71\% relative to the strongest neural baseline. TSP and CVRP experiments evaluate in-distribution performance and generalization across distributions and problem sizes.

Our contributions are:
\begin{enumerate}
  \item We decouple one-step deterministic transition consequences from latent neural matching: the environment computes exact timing and travel effects, the decoder state and mask handle feasibility and capacity, and the resulting decision signal is factorized into a linear comparison channel and a non-linear modulation path.
  \item We realize this decoupling without re-encoding: the consequences enter as a low-dimensional block alongside the cached candidate keys and are scored with one state-conditioned weight shared by all candidates, so the encoder still runs once per instance and its weights remain responsible only for the static node representations.
  \item We evaluate LINC on CVRPTW, CVRP, and TSP with unguided and SGBS-guided decoding, external benchmarks, no-retraining scale transfer, equal-instance continuation controls, and ablations supporting that the decoupled local-consequence interface and step-level summary are central to the gains.
\end{enumerate}

\section{Related Work}

\paragraph{Constructive neural routing solvers.} Pointer Networks introduced sequence-to-sequence construction for combinatorial optimization \citep{vinyals2015pointer}, followed by reinforcement-learning solvers for TSP and VRP \citep{bello2017neural,nazari2018reinforcement}. The Attention Model (AM) \citep{kool2019attention} established the encoder-decoder template that remains the default baseline for routing NCO. POMO \citep{kwon2020pomo} exploited multiple equivalent starting solutions to reduce policy-gradient variance and improve inference through multi-start rollouts. LINC is compatible with this constructive setup: it does not change the MDP or add local search operators, but changes the candidate scoring interface used at each autoregressive step.

\paragraph{Diversity, populations, and guided inference.} A major line of work improves neural solvers by expanding the set of trajectories considered during training or inference. Population policies and winner-take-all objectives encourage complementary strategies \citep{grinsztajn2023winner,hottung2025polynet}, while latent-space policy adaptation searches over policy variants at test time \citep{chalumeau2023combinatorial}. Guided neural decoding methods such as SGBS, DPDP, MDAM, DACT, and NeuroLKH combine learned signals with beam search, dynamic programming, improvement, or LKH-style inference \citep{choo2022simulation,kool2022deep,xin2021multi,ma2021learning,xin2021neurolkh}. These methods answer ``which trajectories should be explored?'' LINC addresses a different question: given a particular partial solution and feasible set, what information should the decoder compare before assigning logits?

\paragraph{Generalization and decoder bottlenecks.} Several recent papers argue that constructive NCO fails out of distribution because the model retains decision-irrelevant information or places too much burden on static embeddings. BQ-NCO formalizes a bisimulation quotient for constructive MDPs, showing how removing irrelevant state distinctions can improve generalization \citep{drakulic2023bq}. LEHD shifts capacity from a light encoder to a heavier decoder for large-scale generalization \citep{luo2023neural}, and ReLD re-examines light-decoder solvers, arguing that static embeddings become too dense as VRP complexity increases \citep{huang2025rethinking}. LINC is closest in motivation to this decoder-bottleneck view, but its mechanism is different: rather than only increasing decoder capacity, it externalizes transition-coupled candidate consequences and routes them through a decoupled comparison-and-modulation interface. The explicit candidate branch computes local consequences, while the summary-modulated comparator learns their state-dependent relevance; unlike BQ-NCO's full MDP quotienting, this separation operates at the single-step scoring interface.

\paragraph{Algorithmic alignment and architectural decoupling.} The concept of algorithmic alignment \citep{xu2020neural} argues that networks whose computational graph mirrors the target algorithm's structure generalize better. This principle has been explored through relational graph networks \citep{battaglia2018relational} and symmetry-aware architectures \citep{bronstein2021geometric}. LINC applies this philosophy to candidate scoring: the known transition supplies local consequences, a shared linear comparator scores them, and feasible-set summaries modulate their state-dependent weighting. This provides an explicit path for deterministic quantities to influence action preferences alongside learned matching.

\paragraph{VRP variants and constraint handling.} Another line focuses on broader VRP coverage, robustness, and complex constraints, including omni-generalizable VRP solvers, robust routing policies, complex-constraint handling, decomposed constraint retrieval, and multi-task mixture-of-experts formulations \citep{zhou2023towards,zhou2024collaboration,bi2024learning,jiang2025droc,zhou2024mvmoe,berto2025rl4co}. These works seek variant-level generality or robustness across problem families. LINC exposes variant-specific local consequences through a shared comparison-and-modulation interface. Its contribution concerns candidate scoring within each construction step.

\section{Problem Setup and Framework}

\subsection{Background}

We follow the standard constructive NCO formulation: a solution is built sequentially as $\xi_{\le t}=(a_1,\dots,a_t)$, with the policy factorizing as $\pi_\theta(\xi\mid \mathcal{I})=\prod_t \pi_\theta(a_t\mid s_t,\mathcal{I})$. TSP, CVRP, and CVRPTW share this autoregressive paradigm but differ in state variables---TSP tracks visited nodes; CVRP adds remaining capacity and depot resets; CVRPTW further adds current time, service times, and customer time windows. We do not alter the MDP definitions or transitions; LINC only changes the decoder-side scoring interface used to assign logits to feasible actions in $\cA_t$.

\subsection{LINC Overview}

At a high level, LINC augments the base decoder logit $b_{t,j}$ with a state-conditioned local consequence comparator:
\begin{equation}
  u_{t,j} = \alpha_t b_{t,j} + v_t^\top \phi_t(j),
  \label{eq:linc}
\end{equation}
where $\phi_t(j)$ is the explicit local consequence representation, $v_t$ is a candidate-shared linear comparator, and $\alpha_t$ gates the base matching path. Since $b_{t,j}=\langle h_t,k_j\rangle$ with keys computed once per instance, Eq.~\eqref{eq:linc} matches an extended query $[\alpha_t h_t; v_t]$ against an extended key $[k_j;\phi_t(j)]$, whose first block is the cached static key and whose second block is recomputed by the environment at each step. The following sections define $\phi_t(j)$, the feasible-set summary used to condition $v_t$ and $\alpha_t$, and the implementation parameterization that realizes this interface.

LINC thus separates two parts of the current-step decision: the learned global matching captured by attention, and the deterministic local consequences of choosing candidate $j$ in state $s_t$, which answer ``if $j$ is selected now, how will the routing state change?'' The environment computes these consequences, and the network learns their state-dependent weights; candidate-level consequences enter a shared linear comparator, while feasible-set summaries modulate the decoder context (Table~\ref{tab:roles}). In the spirit of algorithmic alignment, this reduces the need to infer transition quantities from learned representations, while the policy still learns how to use them in routing decisions.

\begin{table}[t]
  \caption{Functional decoupling of computational responsibilities in LINC.}
  \label{tab:roles}
  \centering
  \small
  \begin{tabular}{ll}
    \toprule
    Responsibility & Realized by \\
    \midrule
    Global policy matching & $b_{t,j}$, the standard context--key matching logit \\
    Exact local consequence computation & Environment / deterministic transition equations \\
    Candidate-level consequence comparison & $\phi_t(j)$ scored by shared linear weight $v_t$ \\
    Feasible-set state & Summary $r_t$ \\
    State-conditioned modulation & $\gamma_t,\beta_t,\alpha_t$ from $[h_t;r_t]$; $v_t=W_\phi^\top\htil_t$ \\
    \bottomrule
  \end{tabular}
\end{table}

\subsection{Candidate-Level Comparison Channel}
\label{sec:conseq}

Let $\cA_t$ be the feasible action set, including the depot when allowed, and $F_t$ its feasible customer subset. For each candidate $j$, the environment computes the local consequences of choosing it in state $s_t$. Using CVRPTW as the running example, let $T_t$ be the current time, $c_t(j)$ the travel time to $j$, $[e_j,l_j]$ its time window, and $q_j$ its service duration. The temporal consequences are
\begin{equation}
  \begin{aligned}
    \arr_t(j) &= T_t + c_t(j), & w_t(j) &= \max\{e_j-\arr_t(j),\,0\},\\
    \dep_t(j) &= \arr_t(j)+w_t(j)+q_j, & \delta_t(j) &= l_j-\arr_t(j)-w_t(j).
  \end{aligned}
  \label{eq:tw}
\end{equation}
Normalized by instance-level scales, these quantities together with task-specific geometric cues form the explicit consequence vector of each candidate customer $j\in F_t$ in state $s_t$:
\begin{equation}
  \phi_t(j) = \big[\hat c_t(j),\, \hat w_t(j),\, \hat\delta_t(j),\, \arrhat_t(j),\, \dephat_t(j),\, \hat\theta_t(j)\big],
  \label{eq:feat}
\end{equation}
where $\hat c_t(j)$ is normalized travel time, $\hat w_t(j)$ is normalized waiting time, $\hat\delta_t(j)$ is normalized time-window slack, $\arrhat_t(j)$ and $\dephat_t(j)$ are normalized arrival and departure times, and $\hat\theta_t(j)$ is the angle between the current node and candidate $j$ as seen from the depot, divided by $\pi$. Together they capture ``what happens immediately if this candidate is chosen.'' For CVRP with remaining capacity $Q_t$, customer demand $d_j$, and depot distance $c_0(j)$, the candidate vector is $\phi_t^{\mathrm{CVRP}}(j)=\big[\hat c_t(j),\, \hat d_t(j),\, \hat c_0(j),\, \hat\theta_t(j)\big]$ with demand ratio $\hat d_t(j)=d_j/\max(Q_t,\epsilon)$; for TSP, $\phi_t^{\mathrm{TSP}}(j)=\big[\hat c_t(j),\, \hat\theta_t^{\mathrm{cent}}(j)\big]$, where $\hat\theta_t^{\mathrm{cent}}(j)$ is the angle relative to the instance centroid divided by $\pi$. When returning to the depot ($j=0$) is feasible, $\phi_t(0)$ is evaluated from the return transition ($\hat c_t(0)$ with zero demand/waiting and zero angle), while capacity, feasibility, and route-reset semantics remain handled by the decoder state and mask.

\subsection{Network Architecture}

\begin{figure}[t]
  \centering
  \includegraphics[width=\textwidth]{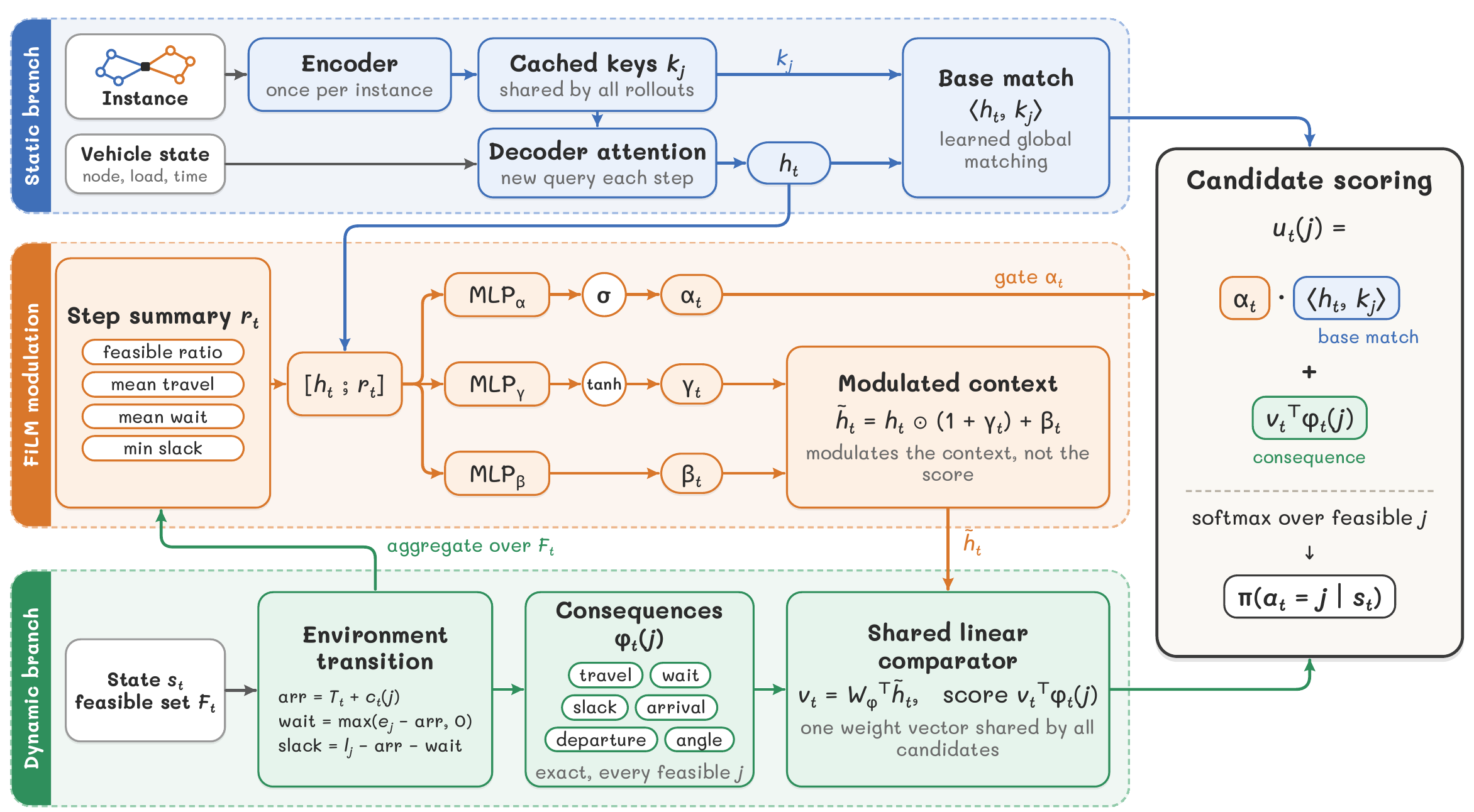}
  \caption{LINC decoder architecture: explicit local-consequence computation is separated from learned state-dependent scoring, with feasible-set summaries modulating the comparator and the retained global matching path.}
  \label{fig:arch}
\end{figure}

Figure~\ref{fig:arch} illustrates the LINC decoder. Instead of compressing all candidate judgments into standard context--key matching, it explicitly introduces a local consequence branch and a feasible-set summary branch before candidate scoring.

Concretely, at step $t$ the decoder produces context $h_t$ and reuses cached candidate keys $k_j$, while the local consequence branch supplies $\phi_t(j)$ from Section~\ref{sec:conseq}. Beyond candidate-level features, we construct a step-level summary from the current feasible customer set $F_t$. For CVRPTW,
\begin{equation}
  r_t = \Big[\rho_t,\ \mathrm{mean}_{j\in F_t}\big(\hat c_t(j)\big),\ \mathrm{mean}_{j\in F_t}\big(\hat w_t(j)\big),\ \min_{j\in F_t}\big(\hat\delta_t(j)\big)\Big],
  \label{eq:summary}
\end{equation}
where $\rho_t=|F_t|/\max(n,1)$ is the feasible-customer ratio. Across tasks, $r_t$ follows the same pattern---feasible-customer ratio, feature means, and the minimum of a key constraint quantity: specifically, CVRP uses $r_t^{\mathrm{CVRP}}=\big[\rho_t,\, \mathrm{mean}_{j\in F_t}(\hat c_t(j)),\, \mathrm{mean}_{j\in F_t}(\hat d_t(j)),\, \min_{j\in F_t}(1-\hat d_t(j))\big]$ with post-service remaining-load ratio $1-\hat d_t(j)$, and TSP uses $r_t^{\mathrm{TSP}}=\big[\rho_t,\, \mathrm{mean}_{j\in F_t}(\hat c_t(j)),\, \min_{j\in F_t}(\hat c_t(j))\big]$.

The context and summary are concatenated and fed through three small MLPs with appropriate activations to produce modulation quantities:
\begin{equation}
  \gamma_t=\tanh\!\big(\mathrm{MLP}_\gamma([h_t;r_t])\big),\quad
  \beta_t=\mathrm{MLP}_\beta([h_t;r_t]),\quad
  \alpha_t=\sigma\!\big(\mathrm{MLP}_\alpha([h_t;r_t])\big),
  \label{eq:mod}
\end{equation}
where $\gamma_t,\beta_t\in\R^d$, $\alpha_t\in(0,1)$. The $\tanh$ nonlinearity constrains $\gamma_t\in(-1,1)$ to prevent over-scaling the context, and $\sigma$ denotes a sigmoid. The modulated context is defined as
\begin{equation}
  \htil_t = h_t\odot(1+\gamma_t)+\beta_t .
  \label{eq:htil}
\end{equation}
The final unnormalized score for candidate $j$ is
\begin{equation}
  u_{t,j} = \alpha_t\langle h_t, k_j\rangle + \big\langle \htil_t,\, W_\phi\,\phi_t(j)\big\rangle .
  \label{eq:score}
\end{equation}
The policy is
\begin{equation}
  \pi_\theta(a_t=j\mid s_t)=\mathrm{softmax}_{j\in\cA_t}(u_{t,j}).
  \label{eq:policy}
\end{equation}
The local-comparator term can be written as $\langle\htil_t, W_\phi\phi_t(j)\rangle = v_t^\top\phi_t(j)$ with $v_t=W_\phi^\top\htil_t$. All candidates share the same state-conditioned weight, so the environment supplies their local effects and the network learns how to compare them. The summary adapts these weights to the current feasible set, while the base attention term retains global routing preferences. Appendices~\ref{app:factor} and~\ref{app:equiv} formalize this factorization and its customer-permutation equivariance.

The consequence branch adds $p_\phi$ features per candidate ($p_\phi=6$ for CVRPTW) and one shared weight vector per step, while the encoder and the candidate keys are still computed once per instance, so the per-step complexity of the base decoder is unchanged. Computing the consequences still adds wall-clock time in our implementation (about $2\times$ in Tables~\ref{tab:cvrptw}--\ref{tab:cvrp}); Appendix~\ref{app:budget} shows that LINC with half the rollout budget, at roughly the same time as PolyNet, retains its advantage on CVRPTW and the external benchmarks.

\subsection{Training Strategy and Credit Assignment}

LINC is trained using REINFORCE \citep{williams1992simple} with sampled $z$-vector rollouts, following the same diversity mechanism as PolyNet \citep{hottung2025polynet}. For each training instance $\mathcal{I}$, the policy samples $K$ solutions $\xi^1,\dots,\xi^K$, each conditioned on a latent vector $z^i$, and is updated with
\begin{equation}
  \nabla_\theta J \approx \frac1K\sum_{i=1}^{K} A_i\,\nabla_\theta\log\pi_\theta(\xi^i\mid \mathcal{I},z^i),
  \qquad \log\pi_\theta(\xi^i\mid \mathcal{I},z^i)=\sum_t\log\pi_\theta(a^i_t\mid s^i_t,\mathcal{I},z^i).
  \label{eq:pg}
\end{equation}
The advantage $A_i$ measures how much better rollout $i$ is than a baseline built from the other rollouts of the same instance, so no learned critic is needed. POMO uses the group-mean baseline, $A_i=\frac1K\sum_j C_j-C_i$ with solution cost $C_i$, and updates every rollout; PolyNet uses hard top-1 credit assignment, which updates only the best rollout of each instance and lets the latent-conditioned policies specialize. For CVRPTW scratch training, we use a soft top-1 advantage scaled to match the group-mean advantage in magnitude:
\begin{equation}
  A_i = (K-1)\big(m_{\mathrm{loo}}(i) - m_{\mathrm{all}}\big),
  \label{eq:soft_adv}
\end{equation}
where $m_{\mathrm{all}} = -\tau\log\big(\frac1K\sum_{j=1}^K e^{g_j}\big)$, $m_{\mathrm{loo}}(i) = -\tau\log\big(\frac1{K-1}\sum_{j\ne i} e^{g_j}\big)$, and $g_i=-(C_i/\kappa)/\tau$ is the temperature-scaled negative cost with a fixed cost scale $\kappa$. Its temperature $\tau$ interpolates between the two regimes: as $\tau\to\infty$ it reduces to the group-mean advantage, and as $\tau\to0$ it approaches top-1 credit assignment (derivation in Appendix~\ref{app:soft}). TSP and CVRP are initialized from converged PolyNet weights and use hard top-1, with training focused on adapting the LINC scoring interface through score morphing.

When training is initialized from weights without the LINC interface, such as POMO or PolyNet, we additionally apply score morphing:
\begin{equation}
  u^{\mathrm{final}}_{t,j} = u^{\mathrm{ref}}_{t,j} + \lambda_{\mathrm{morph}}\big(u_{t,j}-u^{\mathrm{ref}}_{t,j}\big),
  \label{eq:morph}
\end{equation}
where $u^{\mathrm{ref}}_{t,j}$ is the reference policy's score at the same step. $\lambda_{\mathrm{morph}}$ anneals from 0 to 1. TSP and CVRP experiments complete this annealing within the first 100 epochs. For models trained from scratch, $\lambda_{\mathrm{morph}}=1$ and no reference score mixing is used. All neural methods that we retrain (POMO, PolyNet, LINC) use identical data generators and test sets within each task.

\section{Experiments}

\subsection{Problem Definitions and Experimental Setup}

We evaluate three Euclidean routing problems: TSP minimizes closed-tour length; CVRP adds depot returns and vehicle capacity; CVRPTW further adds service times and customer time windows. Following PolyNet \citep{hottung2025polynet} and standard NCO practice, we optimize total travel distance for CVRPTW.

Random test sets follow the Euclidean setup of \citet{kool2019attention}: node coordinates are drawn in the unit square, with training size $n=100$ for TSP and CVRP, and evaluation at $n=100$ in-distribution and $n=150$ cross-scale. External generalization includes TSPLIB50--200, CVRPXML100, and CVRPTW Solomon-100/Homberger-200 \citep{reinelt1991tsplib,uchoa2017new,queiroga2021optimal,solomon1987algorithms,gehring1999parallel}. ``Obj.'' is mean total distance. Benchmark gaps are $G=100\big(\sum_i D_i/\sum_i D_i^{\mathrm{ref}}-1\big)$, using benchmark reference distances (dataset best-known distances for CVRPXML100). Model-relative changes use ratios of mean objectives, with the named comparator as denominator.

We use batch size 128 and rollout width $K=128$ (full hyperparameters in Appendix~\ref{app:hyper}). For TSP and CVRP, LINC continues training from PolyNet checkpoints originally initialized from POMO. For CVRPTW, POMO is trained from scratch on our generator (Appendix~\ref{app:data}); PolyNet continues from POMO following its published procedure \citep{hottung2025polynet}. LINC is trained from scratch on the same generator. Training used a single NVIDIA RTX 5090 32GB GPU; inference timings were measured on a single NVIDIA A100 40GB GPU, the same GPU type used by PolyNet \citep{hottung2025polynet}. Feasibility is enforced at each step by masking infeasible actions; post-hoc verification confirms zero constraint violations in all evaluation runs. Appendix~\ref{app:budget} reports three-seed paired bootstrap checks for small-margin comparisons.

We partition methods into unguided neural constructors (POMO, Poppy, PolyNet, LINC) and guided/search-enhanced methods (SGBS, DPDP, MDAM, DACT, BQ-NCO). For CVRPTW, we use PyVRP \citep{wouda2024pyvrp} for Synthetic-100 and Homberger-200, and the dataset-provided per-instance reference distances for Solomon-100; for TSP and CVRP, Concorde/LKH3 and HGS/LKH3 serve as traditional solver references \citep{applegate2006traveling,helsgaun2017extension,vidal2022hybrid}. Following standard practice \citep{kwon2020pomo,hottung2025polynet,chalumeau2023combinatorial}, we report wall-clock inference time alongside objective gaps. Some baseline results are taken from the tables reported by PolyNet.

\begin{table}[t]
  \caption{Search performance results for CVRPTW benchmarks.}
  \label{tab:cvrptw}
  \centering
  \small
  \setlength{\tabcolsep}{2.3pt}
  \begin{tabular}{llrrrrrrrrr}
    \toprule
    \multirow{2}{*}{Type} & \multirow{2}{*}{Method} & \multicolumn{3}{c}{\makecell{Synthetic-100\\ 10K instances}} & \multicolumn{3}{c}{\makecell{Solomon-100\\ 56 instances}} & \multicolumn{3}{c}{\makecell{Homberger-200\\ 60 instances}} \\
    \cmidrule(lr){3-5}\cmidrule(lr){6-8}\cmidrule(lr){9-11}
    & & Obj. & Gap & Time & Obj. & Gap & Time & Obj. & Gap & Time \\
    \midrule
    Solver & Reference & 1857.063 & -- & 14h & 973.2 & -- & 28m & 2687.7 & -- & 1h \\
    \midrule
    \multirow{4}{*}{Unguided} & POMO (greedy) & 2325.943 & 25.25\% & 12s & 1235.2 & 26.9\% & 0.9s & 4235.323 & 57.58\% & 1s \\
    & POMO (sampling) & 2259.117 & 21.65\% & 40s & 1197.8 & 23.1\% & 1s & 4063.189 & 51.18\% & 3s \\
    & PolyNet & 2016.511 & 8.59\% & 6m & 1107.8 & 13.8\% & 3s & 3712.766 & 38.14\% & 8s \\
    & LINC & \textbf{1920.443} & \textbf{3.41\%} & 12m & \textbf{1044.1} & \textbf{7.3\%} & 6s & \textbf{3083.002} & \textbf{14.71\%} & 15s \\
    \midrule
    \multirow{3}{*}{Guided} & POMO SGBS & 1999.730 & 7.68\% & 9m & 1185.602 & 21.82\% & 11s & 3482.300 & 29.57\% & 1m \\
    & PolyNet SGBS & 1970.843 & 6.13\% & 9m & 1062.620 & 9.18\% & 11s & 3290.331 & 22.42\% & 1m \\
    & LINC SGBS & \textbf{1897.500} & \textbf{2.18\%} & 16m & \textbf{1019.272} & \textbf{4.73\%} & 24s & \textbf{2897.835} & \textbf{7.82\%} & 1m \\
    \bottomrule
  \end{tabular}
\end{table}

\begin{figure}[t]
  \centering
  \includegraphics[width=0.85\textwidth]{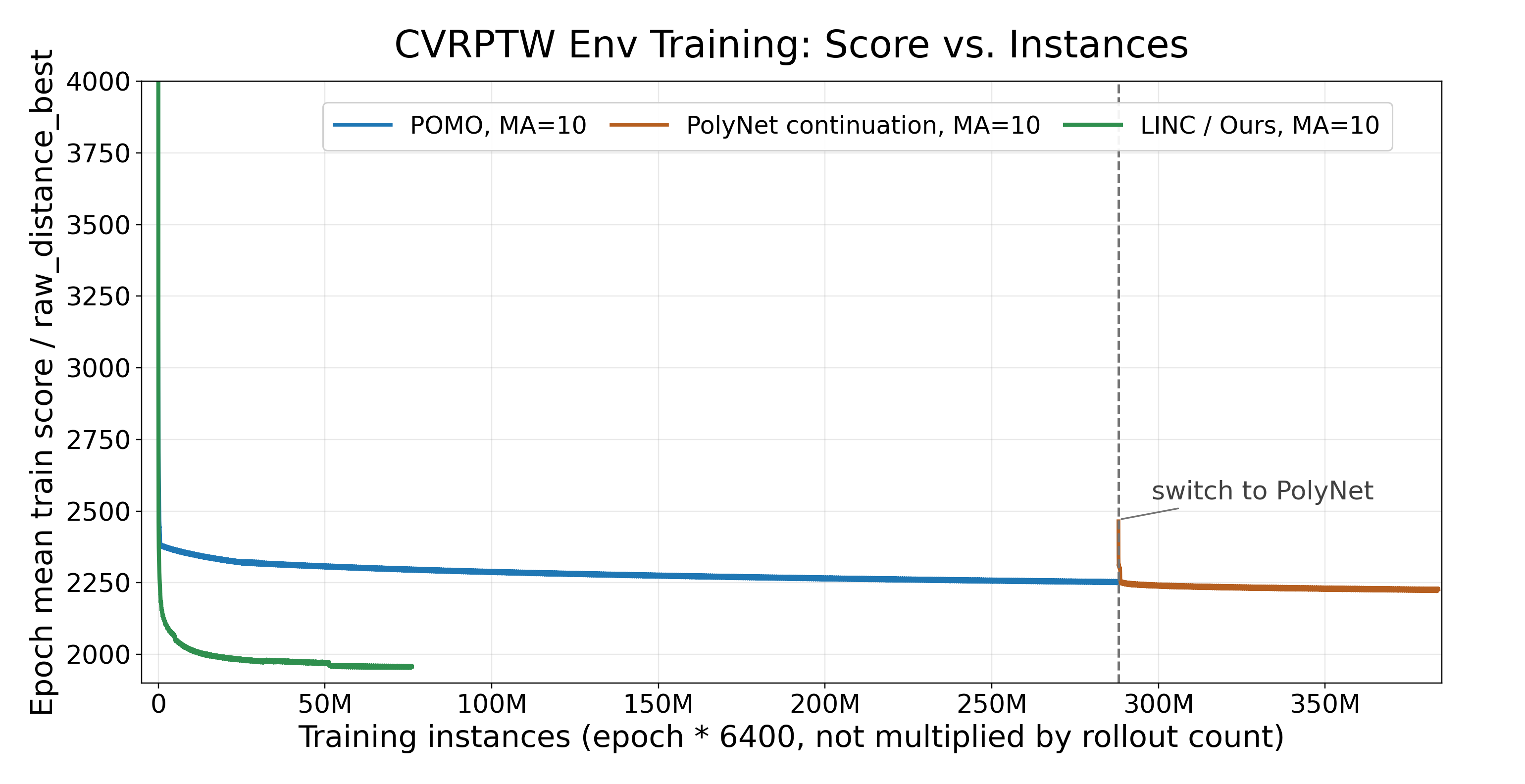}
  \caption{CVRPTW training curves. LINC trains more efficiently, reaching PolyNet's final training score with substantially fewer training instances.}
  \label{fig:train}
\end{figure}

\subsection{CVRPTW Results}

CVRPTW simultaneously couples geometry, capacity, and temporal feasibility, making it the most demanding testbed for explicit local consequence modeling. Table~\ref{tab:cvrptw} reports the benchmark results.

Figure~\ref{fig:train} compares mean best-rollout training distances. LINC reaches PolyNet's final training score with fewer cumulative instances, including POMO pretraining for PolyNet. Gains hold across all three benchmark families in both unguided and SGBS-guided decoding. All neural baselines use the same training generator; Solomon and Homberger are external transfer tests. The largest gains appear on Homberger-200, supporting the usefulness of the explicit consequence interface when transferring to larger constrained-routing instances.

\subsection{TSP and CVRP}

We extend LINC to TSP and CVRP to verify its performance under simpler constraint combinations. Both models are initialized from PolyNet $n100$ weights. For TSP, LINC uses a geometry-conditioned comparator. For CVRP, LINC uses demand-ratio comparator cues and load-after summary cues to sense capacity exhaustion risk.

\begin{table}[t]
  \caption{Search performance results for TSP.}
  \label{tab:tsp}
  \centering
  \small
  \setlength{\tabcolsep}{2.9pt}
  \begin{tabular}{llrrrrrrrrr}
    \toprule
    \multirow{2}{*}{Type} & \multirow{2}{*}{Method} & \multicolumn{3}{c}{\makecell{Test\\ $n_{\mathrm{tr}}=n_{\mathrm{eval}}=100$}} & \multicolumn{3}{c}{\makecell{Generalization\\ $n_{\mathrm{tr}}=100,\ n_{\mathrm{eval}}=150$}} & \multicolumn{3}{c}{\makecell{External benchmark\\ TSPLIB50--200}} \\
    \cmidrule(lr){3-5}\cmidrule(lr){6-8}\cmidrule(lr){9-11}
    & & Obj. & Gap & Time & Obj. & Gap & Time & Obj. & Gap & Time \\
    \midrule
    \multirow{2}{*}{Solver} & Concorde & 7.765 & -- & 82m & 9.346 & -- & 17m & 30478.9 & -- & -- \\
    & LKH3 & 7.765 & 0.000\% & 8h & 9.346 & 0.000\% & 99m & 30478.9 & 0.000\% & -- \\
    \midrule
    \multirow{5}{*}{Unguided} & POMO (greedy) & 7.788 & 0.296\% & 9s & 9.437 & 0.972\% & 2s & 31650.1 & 3.84\% & 15s \\
    & POMO (sampling) & 7.775 & 0.129\% & 38s & 9.393 & 0.500\% & 11s & 31194.6 & 2.35\% & 14s \\
    & Poppy & 7.766 & 0.015\% & 4m & 9.362 & 0.171\% & 1m & -- & -- & -- \\
    & PolyNet & 7.765 & 0.000\% & 4m & 9.352 & 0.064\% & 1m & 31474.2 & 3.27\% & 17s \\
    & LINC & \textbf{7.765} & \textbf{0.000\%} & 8m & \textbf{9.352} & \textbf{0.061\%} & 2m & \textbf{31096.3} & \textbf{2.03\%} & 17s \\
    \midrule
    \multirow{6}{*}{Guided} & DPDP & 7.765 & 0.004\% & 2h & 9.434 & 0.942\% & 44m & -- & -- & -- \\
    & MDAM & 7.781 & 0.208\% & 4h & 9.403 & 0.610\% & 1h & -- & -- & -- \\
    & BQ-NCO & 7.766 & 0.010\% & 25m & 9.351 & 0.057\% & 7m & 30984.6 & 1.66\% & 21s \\
    & POMO SGBS & 7.769 & 0.058\% & 9m & 9.367 & 0.225\% & 8m & 30748.7 & 0.89\% & 2m \\
    & PolyNet SGBS & 7.765 & 0.000\% & 9m & 9.348 & 0.025\% & 2m & 30909.6 & 1.41\% & 2m \\
    & LINC SGBS & \textbf{7.765} & \textbf{0.000\%} & 12m & \textbf{9.348} & \textbf{0.021\%} & 3m & \textbf{30657.3} & \textbf{0.59\%} & 3m \\
    \bottomrule
  \end{tabular}
\end{table}

\begin{table}[t]
  \caption{Search performance results for CVRP.}
  \label{tab:cvrp}
  \centering
  \small
  \setlength{\tabcolsep}{2.9pt}
  \begin{tabular}{llrrrrrrrrr}
    \toprule
    \multirow{2}{*}{Type} & \multirow{2}{*}{Method} & \multicolumn{3}{c}{\makecell{Test\\ $n_{\mathrm{tr}}=n_{\mathrm{eval}}=100$}} & \multicolumn{3}{c}{\makecell{Generalization\\ $n_{\mathrm{tr}}=100,\ n_{\mathrm{eval}}=150$}} & \multicolumn{3}{c}{\makecell{External benchmark\\ CVRPXML100}} \\
    \cmidrule(lr){3-5}\cmidrule(lr){6-8}\cmidrule(lr){9-11}
    & & Obj. & Gap & Time & Obj. & Gap & Time & Obj. & Gap & Time \\
    \midrule
    \multirow{2}{*}{Solver} & HGS & 15.563 & -- & 54h & 19.055 & -- & 9h & 16976.434 & 0.003\% & 7d \\
    & LKH3 & 15.646 & 0.53\% & 6d & 19.222 & 0.88\% & 20h & 17057.239 & 0.479\% & 7d \\
    \midrule
    \multirow{5}{*}{Unguided} & POMO (greedy) & 15.969 & 2.61\% & 11s & 20.054 & 5.24\% & 3s & 18419.3 & 8.50\% & 10s \\
    & POMO (sampling) & 15.754 & 1.23\% & 47s & 19.685 & 3.31\% & 14s & 17889.4 & 5.38\% & 53s \\
    & Poppy & 15.685 & 0.78\% & 5m & 19.578 & 2.74\% & 1m & -- & -- & -- \\
    & PolyNet & 15.640 & 0.50\% & 5m & 19.507 & 2.37\% & 2m & 17632.2 & 3.87\% & 7m \\
    & LINC & \textbf{15.638} & \textbf{0.48\%} & 9m & \textbf{19.468} & \textbf{2.17\%} & 3m & \textbf{17518.8} & \textbf{3.20\%} & 14m \\
    \midrule
    \multirow{7}{*}{Guided} & DACT & 15.747 & 1.18\% & 22h & 19.594 & 2.83\% & 16h & -- & -- & -- \\
    & DPDP & 15.627 & 0.41\% & 23h & \textbf{19.312} & \textbf{1.35\%} & 5h & -- & -- & -- \\
    & MDAM & 15.885 & 2.07\% & 5h & 19.686 & 3.31\% & 1h & -- & -- & -- \\
    & BQ-NCO & 15.806 & 1.56\% & 26m & 19.399 & 1.81\% & 7m & 17947.9 & 5.73\% & 26m \\
    & POMO SGBS & 15.659 & 0.62\% & 10m & 19.426 & 1.95\% & 4m & 17511.0 & 3.15\% & 17m \\
    & PolyNet SGBS & 15.617 & 0.35\% & 12m & 19.358 & 1.59\% & 3m & 17402.5 & 2.51\% & 15m \\
    & LINC SGBS & \textbf{15.615} & \textbf{0.33\%} & 18m & 19.333 & 1.46\% & 5m & \textbf{17327.5} & \textbf{2.07\%} & 21m \\
    \bottomrule
  \end{tabular}
\end{table}

Tables~\ref{tab:tsp} and~\ref{tab:cvrp} report the results. On TSP100, PolyNet and LINC both have a reported gap of 0.000\%; on CVRP100, LINC reduces the gap from 0.50\% to 0.48\%. External-benchmark gaps also decrease: on TSPLIB from 3.27\% to 2.03\% (unguided PolyNet) and 1.41\% to 0.59\% (PolyNet SGBS); on CVRPXML100 from 3.87\% to 3.20\% and 2.51\% to 2.07\%. Appendix~\ref{app:budget} reports paired bootstrap intervals for small-margin TSP/CVRP comparisons and a reduced-budget check ($z=400$) showing that reduced-budget LINC retains the main external/CVRPTW advantage over full-budget PolyNet.

\subsection{No-Retraining Scale Transfer and Equal-Instance Continuation}

\begin{table}[t]
  \caption{No-retraining scale transfer. Mean travel objective (lower is better); $\Delta=100(\mathrm{LINC}/\mathrm{PolyNet}-1)$. CVRP capacities are 100/150 at $n=500/1000$.}
  \label{tab:scale}
  \centering
  \small
  \begin{tabular}{lrrrrr}
    \toprule
    Task & Size & POMO & PolyNet & LINC & $\Delta$ \\
    \midrule
    TSP & 500 & 20.17646 & 19.05102 & \textbf{18.83891} & $-1.11\%$ \\
    TSP & 1{,}000 & 32.50891 & 31.09021 & \textbf{30.33201} & $-2.44\%$ \\
    CVRP & 500 & 54.00508 & 53.24394 & \textbf{41.71045} & $-21.66\%$ \\
    CVRP & 1{,}000 & 155.20651 & 115.86376 & \textbf{61.33192} & $-47.07\%$ \\
    CVRPTW & 400 & 13468.00 & 10656.43 & \textbf{7440.70} & $-30.18\%$ \\
    CVRPTW & 600 & 39464.82 & 25876.28 & \textbf{15615.46} & $-39.65\%$ \\
    CVRPTW & 1{,}000 & 136202.78 & 86869.17 & \textbf{42472.66} & $-51.11\%$ \\
    \bottomrule
  \end{tabular}
\end{table}

We evaluate the $n=100$ checkpoints without retraining on TSP/CVRP at $n=500, 1000$ and Homberger at $n=400, 600, 1000$. TSP and CVRP use 1,000 common instances per size; Homberger uses 60. PolyNet and LINC use $z=8n$ latent codes with greedy construction and eight-fold augmentation; POMO uses its native $n$ greedy starts and eight-fold augmentation. Table~\ref{tab:scale} reports mean travel objectives.

\begin{wrapfigure}{r}{0.45\textwidth}
  \vspace{-12pt}
  \centering
  \includegraphics[width=0.44\textwidth]{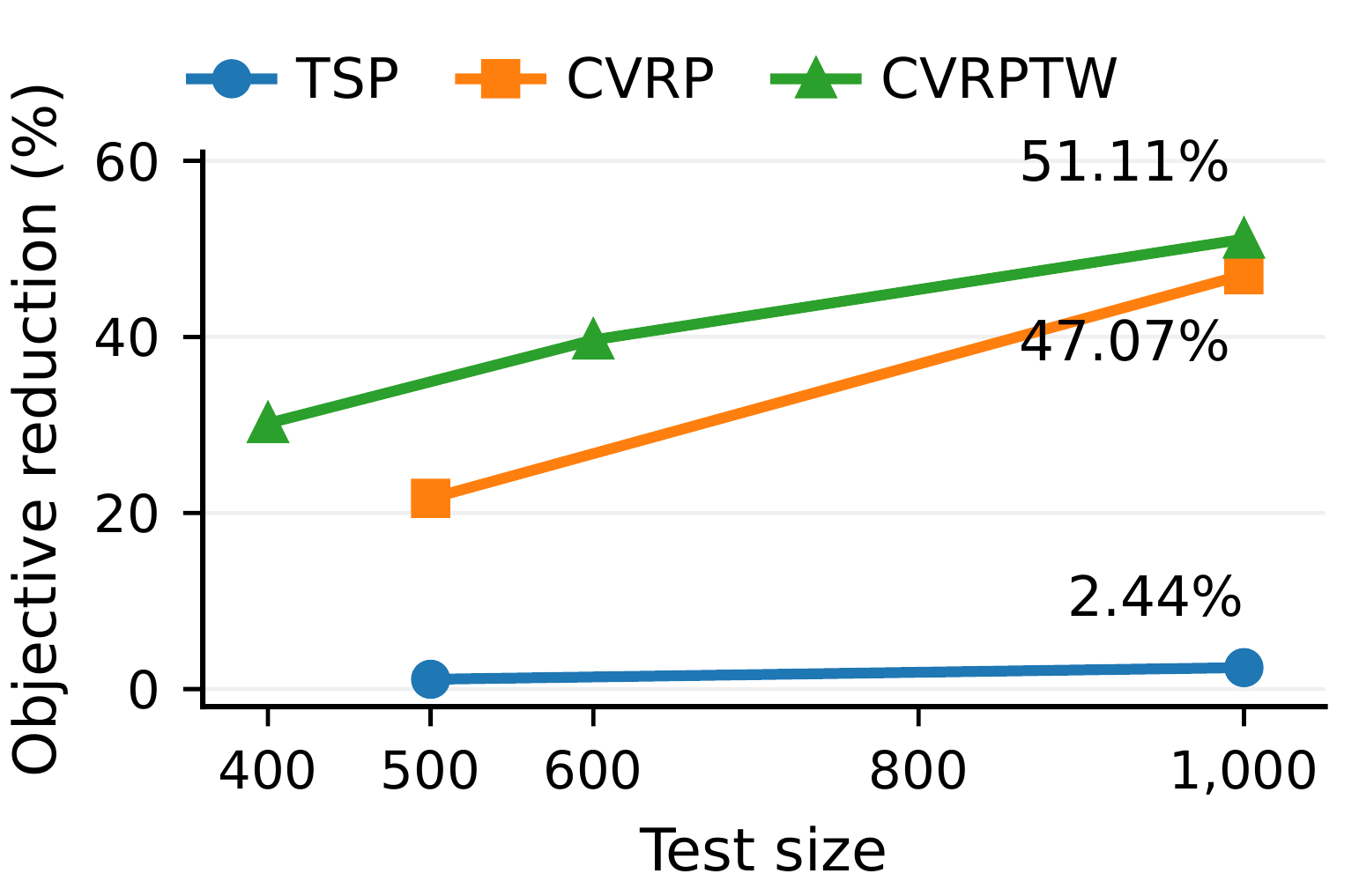}
  \caption{Objective reduction over PolyNet as test size increases. All models are trained at $n=100$.}
  \label{fig:scale}
  \vspace{-8pt}
\end{wrapfigure}
LINC obtains the lowest mean objective at every evaluated large size. Its improvement over PolyNet grows from 1.11\% to 2.44\% on TSP, from 21.66\% to 47.07\% on CVRP, and from 30.18\% to 51.11\% on Homberger (Figure~\ref{fig:scale}). The consequence features are computed by the same equations at every size and need not extrapolate from the training scale, which is consistent with the growing margin.

To separate architectural gains from extra training, we continue PolyNet with its optimizer and scheduler for 2,086,400 additional TSP instances and 9,152,000 CVRP instances, matching LINC's counts. LINC improves on the continued CVRP model in all six settings (Kool100, Kool150, and XML100, each with unguided and SGBS decoding) by 0.047--0.699\%; every paired-bootstrap interval favors LINC. On TSP, the continued PolyNet does not improve on the released checkpoint at TSP100 or TSP150, so additional training alone does not account for LINC's TSP results (Appendix~\ref{app:cont}).

\subsection{Ablation}

\noindent
\begin{minipage}[t]{0.55\textwidth}
\vspace{0pt}
To analyze the sources of LINC's gains, we conduct ablations on a simplified CVRPTW dataset. All variants use identical training and validation sets and the same training budget.

Removing the local consequence interface increases the objective by 2.26\%; replacing it with a naive 2-layer MLP (134$\to$64$\to$1, GELU) increases it by 2.31\%. Removing the step summary increases the objective by 3.15\%. These comparisons support the shared local comparator and step-level modulation.
\end{minipage}\hfill
\begin{minipage}[t]{0.41\textwidth}
\vspace{0pt}
\centering
\footnotesize
\setlength{\tabcolsep}{3pt}
\begin{tabular}{lrr}
  \toprule
  Variant & Obj. & Rel.\ gap \\
  \midrule
  Baseline & 631.65 & 6.66\% \\
  Full LINC & 592.21 & 0.00\% \\
  \makecell[l]{w/o local consequence\\ interface} & 605.60 & 2.26\% \\
  Naive MLP injection & 605.89 & 2.31\% \\
  w/o step summary & 610.85 & 3.15\% \\
  \bottomrule
\end{tabular}
\captionof{table}{Ablation on simplified CVRPTW. Rel.\ gap is relative to Full LINC.}
\label{tab:ablation}
\end{minipage}

\medskip
Together, these ablations support the decoupling: explicit consequences enter a state-conditioned shared comparator, while feasible-set information adapts the current-step scoring. Table~\ref{tab:ablation_full} reports the component ablations.

\subsection{Limitations}

LINC inherits the sequential construction paradigm of POMO and PolyNet: a single sequence is built and split into routes at depot visits. The policy therefore has no explicit mechanism for coordinating vehicles. Once a route returns to the depot it is fixed, and its customers cannot be exchanged with later routes; any coordination must be anticipated implicitly when the earlier routes are built, by leaving suitable customers for vehicles that have not yet been dispatched. This can lead to early commitment and imbalanced routes, and the one-step consequence features do not describe these cross-route effects. Explicit multi-route coordination, for example through parallel route construction or route-level lookahead, is left for future work. In addition, the explicit consequence term can make training harder in some settings: when it dominates the logits early in training, the policy can collapse toward a myopic rule that repeatedly selects the locally cheapest candidate, which reduces exploration and slows the learning of longer-range trade-offs.

\section{Conclusion}

We presented LINC, a decoder that separates what the routing transition already computes from what the neural policy must learn: candidate-level consequences are scored by a state-conditioned shared linear comparator, while global state context flows through nonlinear modulation. The largest gains occur on CVRPTW; TSP and CVRP results also show improved external-benchmark performance. Future work can expose cheap short-horizon consequence summaries through the same interface, moving from one-step comparison toward predictive local comparison without forcing lookahead arithmetic back into the latent matcher. More broadly, these results suggest a decoupling principle for neural decision models: quantities that a known transition, simulator, or constraint process can compute easily should be computed explicitly and given to the network, which then only has to learn how to weigh them.

\begin{ack}
\textbf{Funding in direct support of this work:} This work utilized GPU computing resources supported by the institutional research funds of the corresponding author, Li Wang, at Beijing University of Posts and Telecommunications. No other third-party funding or support was received in direct support of this work during the 36 months prior to submission.

\textbf{Additional revenues related to this work:} The first author gratefully acknowledges support from the 2026 NeurIPS Financial Assistance Program for conference attendance.

\textbf{Competing interests:} The authors declare no competing interests.
\end{ack}

\appendix
\let\oldsection\section
\renewcommand{\section}{\FloatBarrier\oldsection}

\section{Decoupled Local-Consequence Scoring}
\label{app:factor}

At each construction step, LINC separates the deterministic candidate representation $\phi_t(j)$ from its learned state-dependent weighting. The following algebraic factorization makes these computational roles explicit.

\paragraph{State-conditioned linear factorization.} Fix a step with modulated context $\htil_t\in\R^d$, shared projection $W_\phi\in\R^{d\times p_\phi}$, and optional projection bias $a\in\R^d$. Expanding the inner product in the local score
\begin{equation}
  \ell_{t,j} = \big\langle \htil_t,\, W_\phi\phi_t(j)+a\big\rangle
\end{equation}
and transposing the shared projection yields one candidate-shared weight $v_t=W_\phi^\top\htil_t\in\R^{p_\phi}$ and a candidate-independent scalar offset $c_t=\htil_t^\top a$:
\begin{equation}
  \ell_{t,j}=v_t^\top\phi_t(j)+c_t,\qquad \ell_{t,j}-\ell_{t,k}=v_t^\top\big(\phi_t(j)-\phi_t(k)\big),
\end{equation}
where subtracting two candidates cancels $c_t$. Unlike a nonlinear per-candidate scorer over state--candidate pairs, this linear structure summarizes the state-dependent comparison into a single low-dimensional vector $v_t\in\R^{p_\phi}$ shared by all candidates at step $t$, which also directly serves as interpretable marginal preference weights over the consequence features $\phi_t(j)$ (Appendix~\ref{app:diag}, Figure~\ref{fig:weights}).

The complete score combines this explicit local comparison with the global attention preference:
\begin{equation}
  u_{t,j}-u_{t,k} = \alpha_t\,(b_{t,j}-b_{t,k}) + v_t^\top\big(\phi_t(j)-\phi_t(k)\big).
\end{equation}
Thus, transition arithmetic determines the candidate consequences, while the summary-modulated context determines how to weigh them. Because $v_t$ depends nonlinearly on $[h_t;r_t]$, the comparison adapts to changing routing state without recomputing a separate nonlinear candidate embedding.

The released code additionally subtracts the feasible-customer mean from several candidate features before scoring; this adds a common offset to all candidate logits and does not affect the conclusions.

\section{Permutation-Equivariant Consequence Interface}
\label{app:equiv}

\textbf{Proposition B.1} (customer permutation equivariance)\textbf{.} \textit{Let $P$ permute customer labels while leaving the depot fixed. Assume that the candidate-feature computation and preprocessing, candidate keys, base scores, and feasibility mask follow the same row permutation, and that the decoder context and feasible-set summary are unchanged under consistent relabeling of the routing state. For a fixed latent policy code, the LINC score and probability vectors are permutation equivariant.}

\textit{Proof.} Write $\Phi_t$ for the matrix whose rows are $\phi_t(j)^\top$. Invariance of $h_t$ and $r_t$ leaves $\alpha_t$, $\htil_t$, and $v_t$ unchanged. Since $\Phi_t$ and $b_t$ become $P\Phi_t$ and $Pb_t$,
\begin{equation}
  u_t' = \alpha_t P b_t + (P\Phi_t)v_t + c_t\mathbf{1} = P\big(\alpha_t b_t + \Phi_t v_t + c_t\mathbf{1}\big) = P u_t .
\end{equation}
A shared elementwise score transformation, the consistently permuted feasibility mask, and softmax commute with $P$. The probability vector therefore follows the same permutation. \hfill$\square$

\paragraph{Feasible-set conditioning.} The means, minimum slack, and feasible-customer count used in $r_t$ are invariant to customer order. They provide a direct description of the current alternatives to the coefficient-generating network. Candidate rows are evaluated by the shared comparator, while these set-level statistics adapt the comparator and base-matching gate. This preserves the distinction between candidate-level effects and state-level decision context.

\section{Solomon and TSPLIB Supplementary Results}
\label{app:solomon}

This section provides per-instance breakdowns on external benchmarks. Solomon references are the distances stored with the 56-instance dataset. Unguided Solomon results are available as one-decimal per-instance records, so their summary objectives and gaps are reported to one decimal place. The SGBS columns use complete 56-instance verification records (seed 1234, beam width 4, expansion size 4, and 8-fold augmentation; PolyNet and LINC use 128 latent rollouts). Solomon and TSPLIB means and gaps follow the aggregate definition in Section~4.1 and are computed before rounding. Solomon runtimes follow the original NeurIPS evaluation table. TSPLIB tables use 29 EUC\_2D instances with integer-distance objectives.

\begin{table}[!htbp]
  \caption{Per-instance Solomon56 results. Ref.\ contains the dataset-provided reference distances. SGBS means use unrounded verification records; unguided means use the archived one-decimal records. Values are route distances.}
  \label{tab:solomon}
  \centering
  \scriptsize
  \setlength{\tabcolsep}{4.5pt}
  \renewcommand{\arraystretch}{0.92}
  \begin{tabular}{lrrrrrrrr}
    \toprule
    Instance & Ref. & POMO-g & POMO-s & PolyNet & LINC & POMO-SGBS & PolyNet-SGBS & LINC-SGBS \\
    \midrule
    C101 & 827.3 & 934.4 & 947.0 & 843.1 & 828.9 & 869.6 & 828.9 & 828.9 \\
    C102 & 827.3 & 1063.4 & 1073.9 & 953.7 & 831.5 & 1068.9 & 942.7 & 831.7 \\
    C103 & 826.3 & 1139.1 & 1141.4 & 1036.3 & 848.1 & 1127.9 & 963.7 & 838.7 \\
    C104 & 822.9 & 1156.3 & 1130.0 & 957.1 & 861.0 & 1130.7 & 940.1 & 839.1 \\
    C105 & 827.3 & 908.0 & 908.0 & 843.1 & 828.9 & 869.6 & 828.9 & 828.9 \\
    C106 & 827.3 & 1082.1 & 969.0 & 846.5 & 829.7 & 911.6 & 834.8 & 828.9 \\
    C107 & 827.3 & 1004.3 & 907.7 & 881.2 & 832.4 & 879.0 & 831.0 & 828.9 \\
    C108 & 827.3 & 1082.6 & 1059.9 & 843.7 & 835.4 & 1049.6 & 834.8 & 828.9 \\
    C109 & 827.3 & 1034.7 & 1033.9 & 864.3 & 878.1 & 1016.7 & 881.3 & 850.9 \\
    C201 & 589.1 & 861.6 & 823.5 & 766.5 & 646.7 & 821.1 & 642.5 & 614.9 \\
    C202 & 589.1 & 771.7 & 771.7 & 715.1 & 672.1 & 805.3 & 674.9 & 617.7 \\
    C203 & 588.7 & 916.8 & 865.6 & 761.5 & 645.5 & 874.9 & 700.0 & 591.2 \\
    C204 & 588.1 & 895.9 & 870.0 & 759.1 & 637.3 & 836.8 & 728.1 & 644.9 \\
    C205 & 586.4 & 837.9 & 809.2 & 689.2 & 590.5 & 813.9 & 655.3 & 588.9 \\
    C206 & 586.0 & 906.9 & 882.2 & 664.7 & 589.9 & 874.9 & 588.9 & 588.5 \\
    C207 & 585.8 & 771.1 & 760.1 & 658.2 & 604.3 & 736.5 & 610.5 & 588.3 \\
    C208 & 585.8 & 861.0 & 840.5 & 669.9 & 592.6 & 840.3 & 611.1 & 588.3 \\
    R101 & 1637.7 & 1885.7 & 1813.6 & 1794.0 & 1768.5 & 1812.3 & 1722.0 & 1704.0 \\
    R102 & 1466.6 & 1677.4 & 1652.4 & 1612.1 & 1564.4 & 1643.2 & 1577.2 & 1549.6 \\
    R103 & 1208.7 & 1456.2 & 1417.1 & 1390.1 & 1363.0 & 1408.2 & 1341.6 & 1314.9 \\
    R104 & 971.5 & 1198.2 & 1176.5 & 1110.0 & 1060.2 & 1156.5 & 1090.6 & 1034.7 \\
    R105 & 1355.3 & 1640.4 & 1643.0 & 1493.6 & 1486.7 & 1573.2 & 1457.4 & 1419.3 \\
    R106 & 1234.6 & 1554.5 & 1444.8 & 1363.0 & 1334.7 & 1443.8 & 1337.7 & 1335.1 \\
    R107 & 1064.6 & 1320.6 & 1263.2 & 1192.7 & 1178.6 & 1264.6 & 1186.1 & 1134.9 \\
    R108 & 932.1 & 1160.7 & 1127.2 & 1065.8 & 1001.5 & 1135.1 & 1035.5 & 996.6 \\
    R109 & 1146.9 & 1523.4 & 1420.8 & 1302.4 & 1262.4 & 1373.5 & 1274.3 & 1235.4 \\
    R110 & 1068.0 & 1308.2 & 1273.3 & 1187.3 & 1173.1 & 1266.3 & 1153.3 & 1142.8 \\
    R111 & 1048.7 & 1340.2 & 1250.4 & 1200.0 & 1143.0 & 1268.5 & 1179.1 & 1087.9 \\
    R112 & 948.6 & 1151.5 & 1113.8 & 1074.8 & 1025.3 & 1110.4 & 1034.6 & 1009.0 \\
    R201 & 1143.2 & 1343.9 & 1354.5 & 1275.9 & 1237.5 & 1341.7 & 1241.5 & 1206.7 \\
    R202 & 1029.6 & 1195.9 & 1208.1 & 1170.7 & 1104.6 & 1207.5 & 1124.0 & 1089.7 \\
    R203 & 870.8 & 1104.5 & 1072.6 & 987.4 & 937.9 & 1066.0 & 950.4 & 921.8 \\
    R204 & 731.3 & 883.1 & 865.3 & 803.8 & 767.3 & 863.5 & 795.8 & 761.4 \\
    R205 & 949.8 & 1188.9 & 1153.0 & 1077.3 & 1006.9 & 1152.5 & 1033.5 & 991.4 \\
    R206 & 875.9 & 1069.3 & 1038.6 & 997.2 & 933.3 & 1046.5 & 946.0 & 920.2 \\
    R207 & 794.0 & 1004.7 & 966.3 & 918.6 & 857.5 & 984.9 & 859.4 & 842.5 \\
    R208 & 701.0 & 798.1 & 798.1 & 767.4 & 732.2 & 795.2 & 755.8 & 727.2 \\
    R209 & 854.8 & 1076.1 & 1017.6 & 957.2 & 901.3 & 1014.0 & 931.6 & 902.1 \\
    R210 & 900.5 & 1144.5 & 1048.0 & 1021.9 & 964.0 & 1047.1 & 989.4 & 935.6 \\
    R211 & 746.7 & 915.1 & 908.4 & 820.5 & 803.2 & 889.8 & 821.1 & 789.6 \\
    RC101 & 1619.8 & 2050.6 & 2044.3 & 1858.8 & 1794.3 & 2031.1 & 1732.2 & 1734.5 \\
    RC102 & 1457.4 & 1891.3 & 1846.6 & 1691.7 & 1574.6 & 1787.5 & 1608.5 & 1511.5 \\
    RC103 & 1258.0 & 1724.5 & 1611.3 & 1495.3 & 1401.5 & 1611.7 & 1390.6 & 1373.4 \\
    RC104 & 1132.3 & 1409.5 & 1384.5 & 1363.6 & 1248.7 & 1384.0 & 1298.2 & 1236.8 \\
    RC105 & 1513.7 & 1947.4 & 1934.4 & 1772.3 & 1695.8 & 1924.6 & 1637.4 & 1611.6 \\
    RC106 & 1372.7 & 1758.2 & 1706.1 & 1594.3 & 1494.2 & 1670.7 & 1501.8 & 1415.5 \\
    RC107 & 1207.8 & 1661.6 & 1591.1 & 1420.7 & 1360.1 & 1576.7 & 1384.7 & 1294.5 \\
    RC108 & 1114.2 & 1444.5 & 1409.5 & 1351.4 & 1267.6 & 1392.6 & 1277.7 & 1229.6 \\
    RC201 & 1261.8 & 1506.7 & 1527.2 & 1409.8 & 1329.5 & 1497.1 & 1363.9 & 1297.4 \\
    RC202 & 1092.3 & 1382.3 & 1303.8 & 1245.5 & 1142.6 & 1297.0 & 1179.3 & 1131.5 \\
    RC203 & 923.7 & 1164.8 & 1145.7 & 1085.1 & 986.2 & 1140.3 & 1012.0 & 962.0 \\
    RC204 & 783.5 & 989.9 & 981.8 & 918.2 & 826.2 & 957.1 & 877.6 & 803.0 \\
    RC205 & 1154.0 & 1363.9 & 1392.6 & 1287.9 & 1227.4 & 1386.9 & 1285.7 & 1198.4 \\
    RC206 & 1051.1 & 1322.2 & 1214.4 & 1186.4 & 1120.4 & 1199.8 & 1127.2 & 1097.7 \\
    RC207 & 962.9 & 1350.2 & 1150.8 & 1111.3 & 1016.8 & 1135.3 & 1055.0 & 1002.9 \\
    RC208 & 776.1 & 1037.2 & 1012.6 & 909.6 & 825.9 & 1009.7 & 839.5 & 798.8 \\
    \midrule
    Mean & 973.2 & 1235.2 & 1197.8 & 1107.8 & 1044.1 & 1185.6 & 1062.6 & 1019.3 \\
    \bottomrule
  \end{tabular}
\end{table}

\begin{figure}[!htbp]
  \centering
  \includegraphics[width=0.92\textwidth]{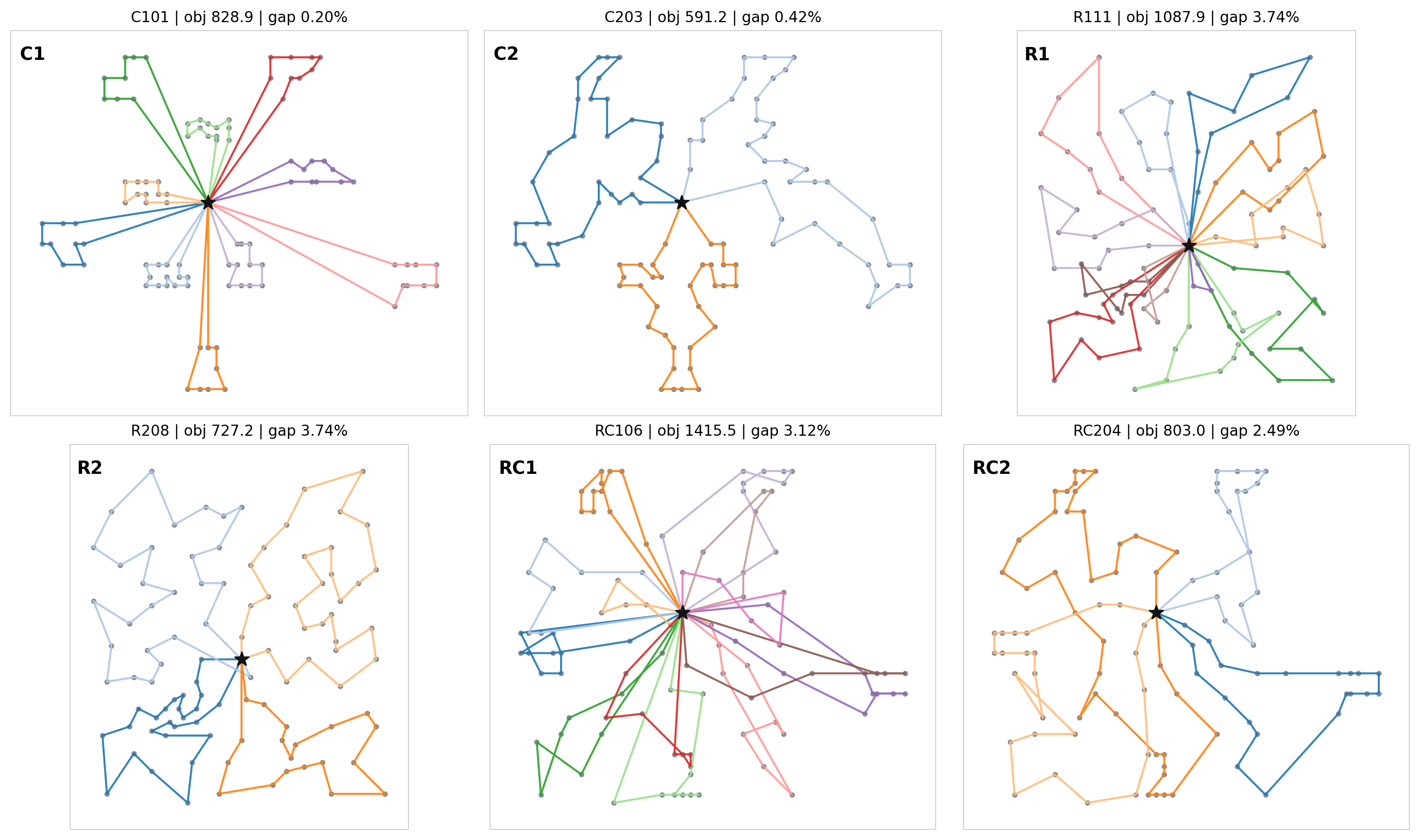}
  \caption{Representative LINC SGBS routes for the six Solomon instance classes.}
  \label{fig:solomon}
\end{figure}

\begin{figure}[!htbp]
  \centering
  \includegraphics[width=0.92\textwidth]{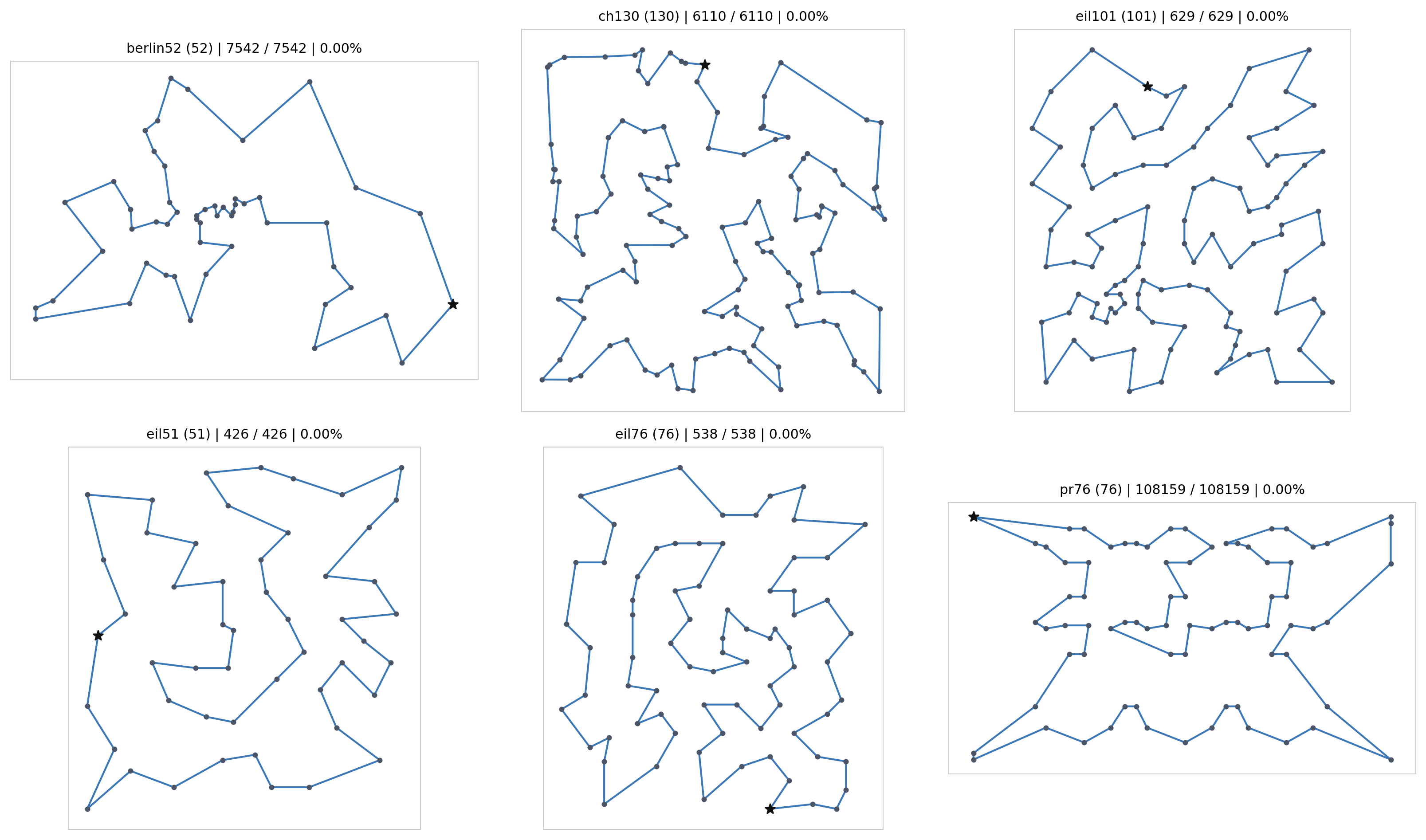}
  \caption{Six TSPLIB50--200 LINC SGBS tours with the lowest gap to the known optimum.}
  \label{fig:tsplib}
\end{figure}

\begin{table}[!htbp]
  \caption{Per-instance TSPLIB50--200 results. Values use EUC\_2D integer distances.}
  \label{tab:tsplib}
  \centering
  \scriptsize
  \setlength{\tabcolsep}{5pt}
  \renewcommand{\arraystretch}{0.92}
  \begin{tabular}{lrrrrrrr}
    \toprule
    Instance & $n$ & Concorde & POMO-g & POMO-s & POMO-SGBS & PolyNet-SGBS & LINC-SGBS \\
    \midrule
    berlin52 & 52 & 7542 & 7543 & 7543 & 7542 & 7542 & 7542 \\
    bier127 & 127 & 118282 & 136654 & 131273 & 122645 & 126686 & 120165 \\
    ch130 & 130 & 6110 & 6136 & 6119 & 6110 & 6110 & 6110 \\
    ch150 & 150 & 6528 & 6565 & 6557 & 6554 & 6528 & 6540 \\
    d198 & 198 & 15780 & 20016 & 18536 & 17543 & 18035 & 17427 \\
    eil101 & 101 & 629 & 632 & 630 & 629 & 629 & 629 \\
    eil51 & 51 & 426 & 428 & 428 & 428 & 426 & 426 \\
    eil76 & 76 & 538 & 541 & 538 & 538 & 538 & 538 \\
    kroA100 & 100 & 21282 & 21466 & 21367 & 21282 & 21282 & 21282 \\
    kroA150 & 150 & 26524 & 27131 & 26707 & 26528 & 26528 & 26528 \\
    kroA200 & 200 & 29368 & 30275 & 29829 & 29494 & 29624 & 29537 \\
    kroB100 & 100 & 22141 & 22399 & 22216 & 22199 & 22141 & 22141 \\
    kroB150 & 150 & 26130 & 26460 & 26460 & 26224 & 26189 & 26132 \\
    kroB200 & 200 & 29437 & 30364 & 29893 & 29536 & 29582 & 29478 \\
    kroC100 & 100 & 20749 & 20987 & 20789 & 20749 & 20749 & 20749 \\
    kroD100 & 100 & 21294 & 21513 & 21472 & 21309 & 21294 & 21374 \\
    kroE100 & 100 & 22068 & 22381 & 22173 & 22120 & 22079 & 22068 \\
    lin105 & 105 & 14379 & 14782 & 14449 & 14379 & 14379 & 14379 \\
    pr107 & 107 & 44303 & 45927 & 44537 & 44351 & 44482 & 44891 \\
    pr124 & 124 & 59030 & 59827 & 59390 & 59076 & 59076 & 59030 \\
    pr136 & 136 & 96772 & 98180 & 97792 & 97420 & 96781 & 97105 \\
    pr144 & 144 & 58537 & 59574 & 58766 & 58588 & 59348 & 58537 \\
    pr152 & 152 & 73682 & 74455 & 74167 & 73690 & 73922 & 73848 \\
    pr76 & 76 & 108159 & 108308 & 108159 & 108159 & 108159 & 108159 \\
    rat195 & 195 & 2323 & 2675 & 2524 & 2407 & 2391 & 2358 \\
    rat99 & 99 & 1211 & 1266 & 1232 & 1215 & 1213 & 1212 \\
    rd100 & 100 & 7910 & 7919 & 7910 & 7910 & 7910 & 7910 \\
    st70 & 70 & 675 & 675 & 675 & 676 & 675 & 676 \\
    u159 & 159 & 42080 & 42773 & 42512 & 42411 & 42080 & 42291 \\
    \midrule
    Mean & -- & 30478.9 & 31650.1 & 31194.6 & 30748.7 & 30909.6 & 30657.3 \\
    \bottomrule
  \end{tabular}
\end{table}

\section{Ablation Supplement: Full Table and Pairwise Comparisons}
\label{app:ablation}

\begin{figure}[!htbp]
  \centering
  \includegraphics[width=0.82\textwidth]{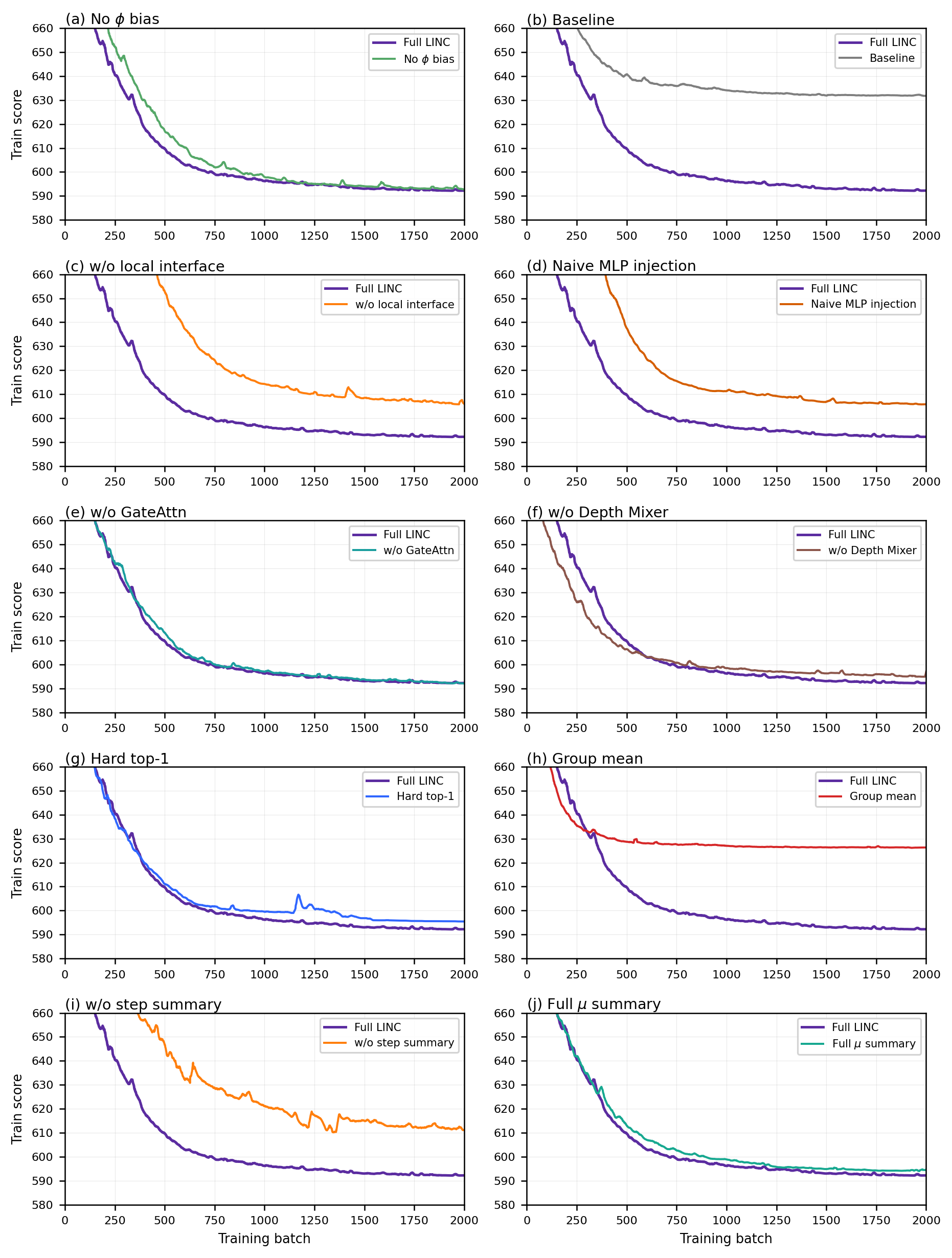}
  \caption{Pairwise training-score comparisons for all ablation variants on the simplified dataset. The full LINC reference curve is shown against each variant over the full training budget.}
  \label{fig:ablation}
\end{figure}

Table~\ref{tab:ablation_full} reports the full ablation table including auxiliary encoder components. Figure~\ref{fig:ablation} provides pairwise training-score comparisons for the component ablations.

\begin{table}[!htbp]
  \caption{Full ablation including auxiliary encoder components. Rel.\ gap is measured relative to Full LINC within the same simplified CVRPTW ablation setting.}
  \label{tab:ablation_full}
  \centering
  \small
  \begin{tabular}{lcccccr}
    \toprule
    Variant & Local cons. & GateAttn & Depth Mixer & Credit & Obj. & Rel.\ gap \\
    \midrule
    Baseline & N & N & N & original & 631.65 & 6.66\% \\
    Full LINC & Y & Y & Y & soft top-1 & 592.21 & 0.00\% \\
    w/o local cons.\ interface & N & Y & Y & soft top-1 & 605.60 & 2.26\% \\
    Naive MLP injection & MLP only & Y & Y & soft top-1 & 605.89 & 2.31\% \\
    w/o GateAttn & Y & N & Y & soft top-1 & 592.06 & $-0.03\%$ \\
    w/o Depth Mixer & Y & Y & N & soft top-1 & 600.73 & 1.44\% \\
    w/o soft top-1 & Y & Y & Y & hard top-1 & 595.59 & 0.57\% \\
    group\_mean & Y & Y & Y & group\_mean & 626.37 & 5.77\% \\
    w/o step summary & Y & Y & Y & soft top-1 & 610.85 & 3.15\% \\
    Full mean summary & Y & Y & Y & soft top-1 & 594.91 & 0.46\% \\
    \bottomrule
  \end{tabular}
\end{table}

\section{Budget Sensitivity and Statistical Checks}
\label{app:budget}

\paragraph{Reduced-budget decoding ($z=400$).} Table~\ref{tab:budget} evaluates LINC with the latent/sample budget reduced from the main setting $z=800$ to $z=400$ while increasing batch size where memory permits. The first five rows are random or in-distribution checks; the remaining rows are external benchmarks. The table reports PolyNet $z=800$, LINC $z=400$, and LINC $z=800$ side by side so that both quality and wall-clock time can be compared directly. This comparison measures the effect of rollout budget on solution quality and inference time.

\begin{table}[!htbp]
  \caption{Reduced-budget check for LINC. All metrics are objective values (lower is better). PolyNet and LINC main ($z=800$) values are from the main result tables; wall-clock times are listed next to each objective for direct comparison.}
  \label{tab:budget}
  \centering
  \small
  \begin{tabular}{llrrrrrr}
    \toprule
    \multirow{2}{*}{Role} & \multirow{2}{*}{Dataset} & \multicolumn{2}{c}{PolyNet $z=800$} & \multicolumn{2}{c}{LINC $z=400$} & \multicolumn{2}{c}{LINC $z=800$} \\
    \cmidrule(lr){3-4}\cmidrule(lr){5-6}\cmidrule(lr){7-8}
    & & Obj. & Time & Obj. & Time & Obj. & Time \\
    \midrule
    \multirow{5}{*}{Random / in-dist.} & TSP100 & 7.765 & 4m & 7.765 & 4m & 7.765 & 8m \\
    & TSP150 & 9.352 & 1m & 9.353 & 1m & 9.352 & 2m \\
    & CVRP100 & 15.640 & 5m & 15.646 & 5m & 15.638 & 9m \\
    & CVRP150 & 19.507 & 2m & 19.483 & 1m & 19.468 & 3m \\
    & CVRPTW synthetic100 & 2016.5 & 6m & 1924.6 & 6m & 1920.4 & 12m \\
    \midrule
    \multirow{4}{*}{External} & CVRPXML100 & 17632.2 & 7m & 17551.4 & 8m & 17518.8 & 14m \\
    & Solomon56 & 1107.8 & 3s & 1049.0 & 3s & 1044.1 & 6s \\
    & Homberger200 & 3712.8 & 8s & 3103.5 & 8s & 3083.0 & 15s \\
    & TSPLIB50--200 & 31474.2 & 17s & 31253.4 & 12s & 31096.3 & 17s \\
    \bottomrule
  \end{tabular}
\end{table}

\begin{figure}[!htbp]
  \centering
  \includegraphics[width=\textwidth]{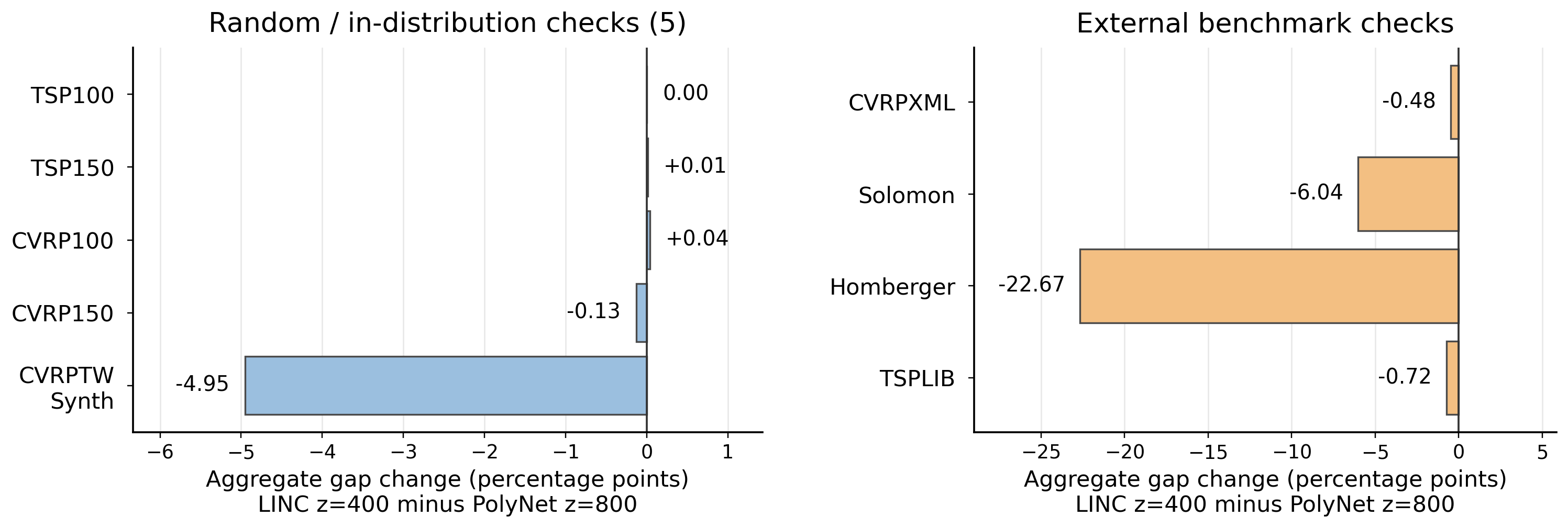}
  \caption{Reduced-rollout sensitivity. Both panels show the aggregate gap difference $G_{\mathrm{LINC},z=400}-G_{\mathrm{PolyNet},z=800}$ in percentage points, computed from the mean objectives in Table~\ref{tab:budget} and the reference distances of the main benchmarks. Negative values favor LINC.}
  \label{fig:budget}
\end{figure}

\paragraph{Paired bootstrap on small-margin random benchmarks.} For random TSP/CVRP benchmarks, we report paired bootstrap intervals over test instances with three pre-specified evaluation seeds (1234, 42, 2026) under fixed checkpoints. These intervals capture instance-level variability of the LINC--PolyNet difference and confirm that the improvement is consistent across seeds. All intervals are negative (LINC lower cost), and none cross zero.

\begin{table}[!htbp]
  \caption{Paired bootstrap checks for small residual-gap comparisons. The reported difference is LINC minus PolyNet; lower is better, so a negative interval favors LINC.}
  \label{tab:bootstrap}
  \centering
  \small
  \begin{tabular}{lrrrrr}
    \toprule
    Benchmark & Seed & LINC & PolyNet & Diff. & 95\% CI \\
    \midrule
    TSP random150 & 1234 & 9.3518 & 9.3528 & $-0.00104$ & $[-0.00139, -0.00070]$ \\
    TSP random150 & 42 & 9.3516 & 9.3527 & $-0.00105$ & $[-0.00142, -0.00068]$ \\
    TSP random150 & 2026 & 9.3519 & 9.3527 & $-0.00082$ & $[-0.00118, -0.00047]$ \\
    CVRP random100 & 1234 & 15.6380 & 15.6402 & $-0.00221$ & $[-0.00286, -0.00154]$ \\
    CVRP random100 & 42 & 15.6379 & 15.6401 & $-0.00219$ & $[-0.00284, -0.00156]$ \\
    CVRP random100 & 2026 & 15.6380 & 15.6400 & $-0.00193$ & $[-0.00257, -0.00130]$ \\
    \bottomrule
  \end{tabular}
\end{table}

\section{Training Hyperparameters}
\label{app:hyper}

All tasks use Adam optimizer ($\beta_1=0.9$, $\beta_2=0.999$, weight decay $10^{-6}$) with amp mixed precision, learning rate $10^{-4}$, batch size 128, and rollout width $K=128$. Each epoch contains 6400 training instances. CVRPTW uses the matched soft top-1 advantage defined in Appendix~\ref{app:soft}. TSP and CVRP anneal $\lambda_{\mathrm{morph}}$ from 0 to 1 over the first 100 epochs. LINC is trained for approximately 300 (TSP), 1400 (CVRP), and 12000 (CVRPTW) epochs respectively. Training uses a single NVIDIA RTX 5090 32GB GPU; reported inference timings use a single A100 40GB GPU.

\section{Data Generation Details}
\label{app:data}

Existing NCO training distributions for CVRPTW can be narrow---restricted to limited spatial layouts, loose time windows, or nearly uniform constraint densities---which makes external benchmark transfer difficult. We therefore design a procedural generator that covers a broader instance envelope through explicit rule-based sampling over spatial morphology, temporal coupling, and constraint intensity. The distribution spans loosely constrained to tightly coupled instances. For each instance, we first sample latent variables
\begin{equation}
  z=\{S, H, \nu, \eta, K, \pi^{\mathrm{space}}, \pi^{\mathrm{width}}, \pi^{\mathrm{phase}}, r_{\mathrm{con}}\},
\end{equation}
where $S$ is the spatial scale, $H$ the time horizon ratio, $\nu$ the service time ratio, $\eta$ the distance-to-time scaling coefficient, $K$ the number of clusters, $\pi^{\mathrm{space}}$, $\pi^{\mathrm{width}}$, and $\pi^{\mathrm{phase}}$ are mixture weights for spatial, time-window width, and phase components, respectively, and $r_{\mathrm{con}}$ is the proportion of constrained nodes. The time horizon endpoint and service time are defined as
\begin{equation}
  T_{\max}=HS,\qquad s=\nu S .
\end{equation}
On the normalized plane $[0,1]^2$, customer locations are generated by mixing four spatial component types: clustered Gaussian components, uniform components, banded corridor components, and outlier components. Let $\tilde x_j$ be the normalized position; the absolute coordinate is
\begin{equation}
  x_j = S\tilde x_j .
\end{equation}
Outlier components are explicitly constrained to be far from the depot and main cluster centers, ensuring that the training envelope includes long-distance, low-density, and non-uniform structures.

For time-window generation, we first compute the feasible upper and lower bounds for each node based on travel time from the depot. Let the depot coordinate be $x_0$; then
\begin{equation}
  \Delta_j=\eta\|x_j-x_0\|_2,
\end{equation}
\begin{equation}
  L_j=\Delta_j,\qquad U_j=T_{\max}-\Delta_j-s .
\end{equation}
Here $L_j$ ensures the vehicle can reach the customer from the depot; $U_j$ ensures sufficient time remains to return to the depot after service.

We then construct normalized phase variables for each node. Let $\psi^{\mathrm{cluster}}_j,\psi^{\mathrm{radial}}_j,\psi^{\mathrm{angular}}_j,\psi^{\mathrm{rand}}_j$ denote cluster, radial, angular, and random phases, and $\pi^{\mathrm{phase}}$ be the 4-dimensional weight vector:
\begin{equation}
  p_j=\mathrm{clip}\Big(\sum_{q\in\{\mathrm{cluster},\mathrm{radial},\mathrm{angular},\mathrm{rand}\}}\pi^{\mathrm{phase}}_q\psi^q_j+\epsilon_j,\ 0,\ 1\Big),
\end{equation}
with $\epsilon_j\sim\mathcal N(0,\sigma_p^2)$. Each node is marked as constrained with probability $r_{\mathrm{con}}$, otherwise unconstrained.

For unconstrained nodes, the time window is set to a global wide window or approximately global wide window. For constrained nodes, we use the phase variable to determine the window center:
\begin{equation}
  m_j=L_j+p_j(U_j-L_j),
\end{equation}
then sample from narrow, medium, and loose Beta width distributions:
\begin{equation}
  \omega_j\sim\sum_{r\in\{\mathrm{narrow},\mathrm{medium},\mathrm{loose}\}}\pi^{\mathrm{width}}_r\,\mathrm{Beta}(a_r,b_r),
\end{equation}
with window width
\begin{equation}
  W_j=\max\{w_{\min},\ \omega_j(U_j-L_j)\}.
\end{equation}
The final time window is
\begin{equation}
  e_j=\max\{L_j,\ m_j-\tfrac12W_j\},\qquad l_j=\min\{U_j,\ m_j+\tfrac12W_j\}.
\end{equation}
This generator thus creates a continuously controllable instance envelope across three dimensions---spatial morphology, temporal coupling structure, and constraint intensity---to cover heterogeneous CVRPTW characteristics.

\section{Encoder Auxiliary Refinements}
\label{app:encoder}

The encoder uses two lightweight auxiliary components inherited from the base architecture:

\paragraph{GateAttn.} Multi-head attention outputs are modulated by head-level gates $g^{(\ell)}_{v,h}=2\sigma(W_g z^{(\ell-1)}_v+b_g)_h$, producing gated outputs $\tilde o^{(\ell)}_{v,h}=(1-\lambda_{\mathrm{gate}})o^{(\ell)}_{v,h}+\lambda_{\mathrm{gate}}g^{(\ell)}_{v,h}o^{(\ell)}_{v,h}$. Gates centered at 1 allow both suppression and enhancement.

\paragraph{Depth Mixer.} Representation history across encoder layers is aggregated via learned depth queries $q^{(\ell)}$, computing attention weights $\omega^{(\ell)}_{n,m}\propto\exp\langle\mathrm{Norm}(v^{(m)}_n),q^{(\ell)}\rangle$ over past layer outputs, with aggregated input $\hat h^{(\ell)}_n=\sum_{m<\ell}\omega^{(\ell)}_{n,m}\mathrm{Norm}(v^{(m)}_n)$.

Table~\ref{tab:ablation_full} reports the encoder ablations alongside the decoder ablations. Removing GateAttn changes the objective by $-0.03\%$; removing Depth Mixer increases it by 1.44\%. The corresponding increases for removing the local-consequence interface and step-level summary are 2.26\% and 3.15\%.

\section{Matched Soft Top-1 Advantage Derivation}
\label{app:soft}

Let $\hat C_i=C_i/\kappa$ be scaled costs, $g_i=-\hat C_i/\tau$ be scaled logits, $m_{\mathrm{all}}=-\tau\log\big(\frac1K\sum_j e^{g_j}\big)$, $m_{\mathrm{loo}}(i)=-\tau\log\big(\frac1{K-1}\sum_{j\ne i}e^{g_j}\big)$. As $\tau\to\infty$, $g_j\to0$, and a Taylor expansion $e^{g_j}\approx1+g_j$ gives:
\begin{equation}
  m_{\mathrm{all}}\approx-\tau\cdot\frac1K\sum_j g_j=\frac1K\sum_j\hat C_j,
\end{equation}
\begin{equation}
  m_{\mathrm{loo}}(i)\approx-\tau\cdot\frac1{K-1}\sum_{j\ne i}g_j=\frac1{K-1}\sum_{j\ne i}\hat C_j .
\end{equation}
Therefore
\begin{equation}
  A_i=(K-1)\big(m_{\mathrm{loo}}(i)-m_{\mathrm{all}}\big)\approx(K-1)\Big(\frac1{K-1}\sum_{j\ne i}\hat C_j-\frac1K\sum_j\hat C_j\Big)=\frac1K\sum_j\hat C_j-\hat C_i,
\end{equation}
which recovers the POMO-style advantage with a group-mean baseline (up to a constant factor $\kappa^{-1}$). The $(K-1)$ factor ensures that the advantage maintains the correct scale at finite $\tau$: without it, the softened advantage magnitude decays as $K$ grows, compressing early training signals.

\section{Mechanism Diagnostics}
\label{app:diag}

We include lightweight diagnostics on the trained CVRPTW checkpoint under the Solomon56 evaluation protocol, inspecting whether the implemented candidate scorer behaves according to the mechanism assumed in the main text.

\begin{figure}[!htbp]
  \centering
  \includegraphics[width=0.82\textwidth]{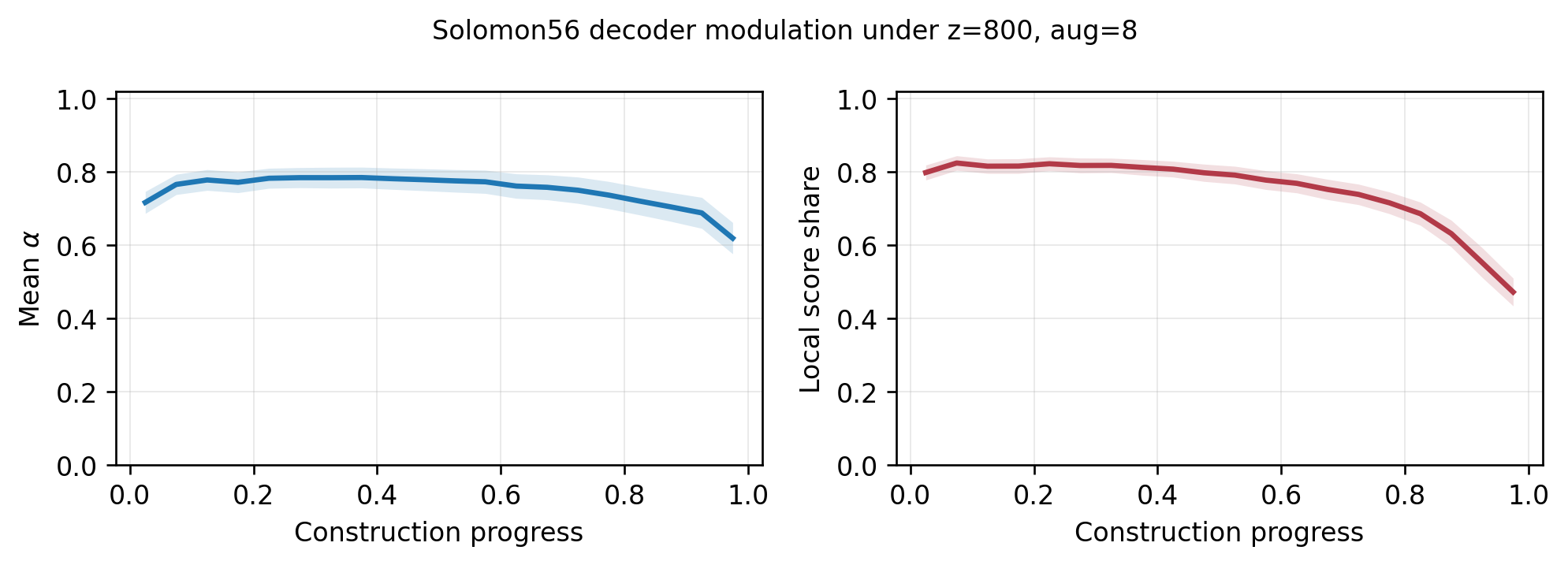}
  \caption{Summary modulation diagnostics on Solomon56. The learned gate and local-score share vary with construction progress.}
  \label{fig:modulation}
\end{figure}

\paragraph{Adaptive modulation.} Figure~\ref{fig:modulation} shows that the learned summary gate $\alpha_t$ remains active through early and middle decoding, while the local-score share decreases near the end of the route when fewer feasible choices remain. The shared local comparator scores candidate-specific consequences, and the summary branch calibrates their influence at each step.

\begin{figure}[!htbp]
  \centering
  \includegraphics[width=0.85\textwidth]{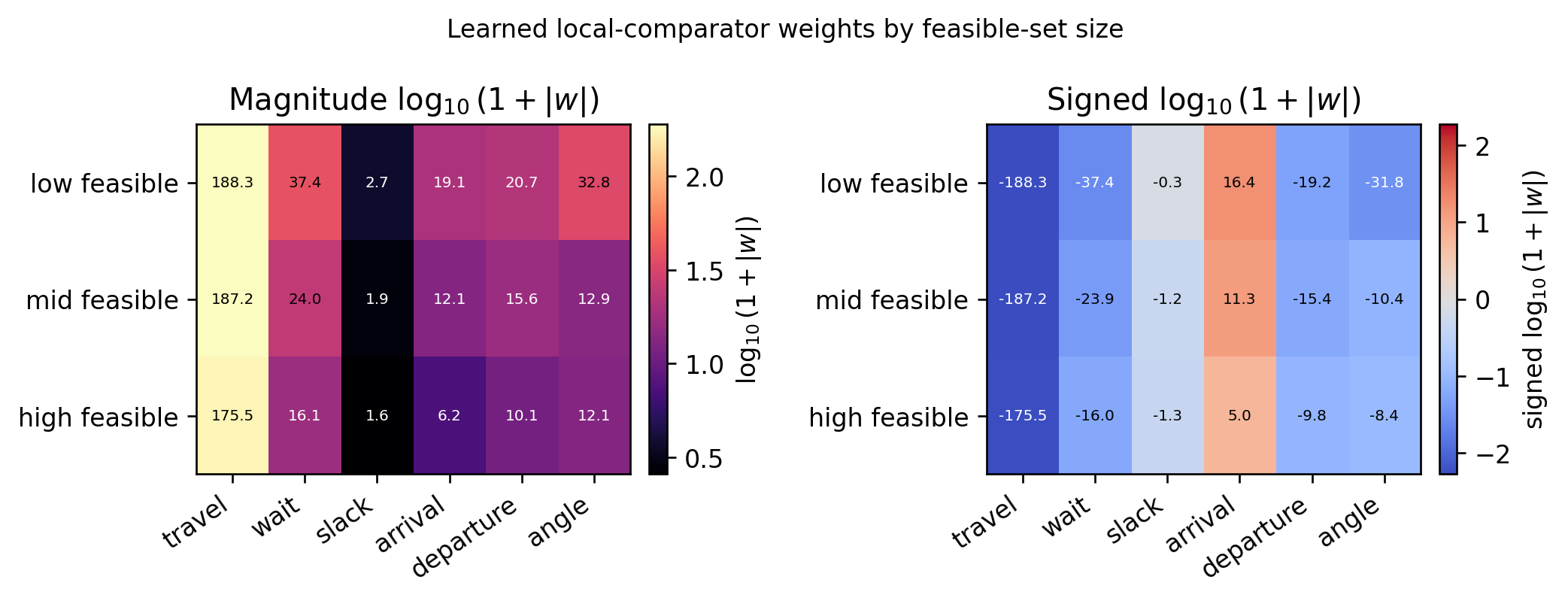}
  \caption{Local-comparator feature weights grouped by feasible-set size. Temporal features gain relative importance when fewer customers remain feasible.}
  \label{fig:weights}
\end{figure}

\paragraph{Feature roles.} Figure~\ref{fig:weights} groups local-comparator weights by feasible-set size. Travel remains the dominant local consequence, while wait, arrival, departure, and angular terms become stronger when the feasible set is small. This matches the CVRPTW intuition that the model first avoids unnecessary travel, then increasingly values temporal urgency as feasible options disappear.

\section{Equal-Instance Continuation Controls}
\label{app:cont}

To test whether additional training alone reproduces the improvement over the released PolyNet checkpoints, we resume PolyNet with its Adam and scheduler states, keeping all parameters trainable. Each epoch generates 6,400 new instances from the same task distribution. TSP adds 326 epochs (epoch 200 to 526; 2,086,400 instances); CVRP adds 1,430 epochs (epoch 300 to 1,730; 9,152,000 instances). These match the additional instance counts used for the respective LINC adaptations.

\paragraph{CVRP.} Table~\ref{tab:cont} compares the original PolyNet, continued PolyNet, and LINC. Kool100/Kool150 use the per-instance average over the three shared evaluation seeds 42, 1234, and 2026; XML100 uses seed 1234. There are 10,000 Kool100/XML100 instances and 1,000 Kool150 instances. Unguided decoding uses $z=800$, augmentation eight, and sampling on Kool100/XML100 but greedy construction on Kool150. SGBS uses $z=128$, beam four, expansion four, and augmentation eight. Models share test instances and decoding budgets within each setting.

\begin{table}[!htbp]
  \caption{CVRP equal-instance continuation. Means are objectives; intervals are for LINC minus continued PolyNet.}
  \label{tab:cont}
  \centering
  \small
  \begin{tabular}{llrrrr}
    \toprule
    Dataset & Decoder & Original & Continued & LINC & Paired 95\% CI \\
    \midrule
    Kool100 & Sampling & 15.63994 & 15.64991 & 15.63807 & $[-0.01244, -0.01122]$ \\
    Kool100 & SGBS & 15.61683 & 15.62256 & 15.61522 & $[-0.00794, -0.00672]$ \\
    Kool150 & Greedy & 19.50803 & 19.50408 & 19.46675 & $[-0.04308, -0.03170]$ \\
    Kool150 & SGBS & 19.35675 & 19.35179 & 19.33228 & $[-0.02374, -0.01524]$ \\
    XML100 & Sampling & 17617.77 & 17642.86 & 17519.47 & $[-128.62127, -118.14187]$ \\
    XML100 & SGBS & 17401.57 & 17403.70 & 17327.82 & $[-79.52432, -72.19579]$ \\
    \bottomrule
  \end{tabular}
\end{table}

All six LINC-minus-continued intervals exclude zero in favor of LINC. They use 20,000 paired bootstrap resamples; two-sided sign tests are Holm-adjusted across the six settings within each model pair, with largest adjusted $p=1.54\times10^{-14}$ for this comparison. Equal-instance continuation does not reproduce LINC's gains.

\paragraph{TSP.} On TSP100, identical latent vectors and reset-identical action random streams give 7.764600 for the released model and 7.764626 for the continued model with eight-fold augmentation. The continued-minus-original difference is $2.56\times10^{-5}$, with paired 95\% interval $[1.34, 3.82]\times10^{-5}$. On TSP150, the corresponding differences are 0.001741 ($[0.001334, 0.002140]$) under unguided sampling and 0.000515 ($[0.000171, 0.000856]$) under SGBS. TSPLIB continuation changes are inconclusive: $-18.72$ ($[-71.69, 27.00]$) for sampling and $-103.45$ ($[-450.42, 97.93]$) for SGBS. These TSP intervals use 10,000 paired bootstrap resamples and compare the two PolyNet checkpoints.

\section{Large-Scale Transfer Results}
\label{app:large}

Table~\ref{tab:scale} uses frozen $n=100$ checkpoints. Random TSP and CVRP each contain 1,000 common instances per size. CVRP demands are drawn from integers 1 to 9, with capacities 100 and 150 at $n=500$ and $n=1000$. Homberger has 60 instances each at $n=400,600,1000$. Reported values are mean objectives on shared instances.

\paragraph{Decoding budget.} POMO uses $n$ native greedy starts per augmentation. PolyNet and LINC use $8n$ latent codes per augmentation, also with greedy construction. All use eight-fold augmentation. PolyNet and LINC have matched population counts at each test size.

\begin{table}[!htbp]
  \caption{Mean number of vehicles on the large Homberger tests. Travel distance is the optimized objective.}
  \label{tab:vehicles}
  \centering
  \small
  \begin{tabular}{rrrr}
    \toprule
    Size & POMO & PolyNet & LINC \\
    \midrule
    400 & 62.60 & 41.12 & 28.75 \\
    600 & 145.20 & 74.63 & 42.55 \\
    1{,}000 & 298.72 & 155.88 & 68.83 \\
    \bottomrule
  \end{tabular}
\end{table}

\clearpage
\renewcommand{\answerNA}[1][]{\textcolor{gray}{[N/A] #1}}

\section*{NeurIPS Paper Checklist}

\begin{enumerate}

\item {\bf Claims}
  \item[] Question: Do the main claims made in the abstract and introduction accurately reflect the paper's contributions and scope?
  \item[] Answer: \answerYes{}
  \item[] Justification: The abstract and introduction clearly state the proposed LINC architecture, its local-consequence interface, and the evaluation scope over TSP, CVRP, and CVRPTW; these claims are substantiated by the method and experimental sections.

\item {\bf Limitations}
  \item[] Question: Does the paper discuss the limitations of the work performed by the authors?
  \item[] Answer: \answerYes{}
  \item[] Justification: Section~4.6 discusses limitations inherited from autoregressive POMO/PolyNet-style construction, including early commitment, route imbalance, and incomplete multi-route coordination, as well as the risk that an overly strong LINC interface degenerates into greedy selection during training.
  \item[] Guidelines:
  \begin{itemize}
    \item The answer \answerNA{} means that the paper has no limitation while the answer \answerNo{} means that the paper has limitations, but those are not discussed in the paper.
    \item The authors are encouraged to create a separate ``Limitations'' section in their paper.
    \item The paper should point out any strong assumptions and how robust the results are to violations of these assumptions.
    \item The authors should reflect on the scope of the claims made, e.g., if the approach was only tested on a few datasets or with a few runs.
    \item The authors should reflect on the factors that influence the performance of the approach.
    \item The authors should discuss the computational efficiency of the proposed algorithms and how they scale with dataset size.
    \item If applicable, the authors should discuss possible limitations of their approach to address problems of privacy and fairness.
    \item Reviewers will be specifically instructed to not penalize honesty concerning limitations.
  \end{itemize}

\item {\bf Theory assumptions and proofs}
  \item[] Question: For each theoretical result, does the paper provide the full set of assumptions and a complete (and correct) proof?
  \item[] Answer: \answerYes{}
  \item[] Justification: Appendices~A--B provide the complete algebraic factorization derivation and customer-permutation-equivariance proposition used in the paper. Their assumptions and complete derivations/proofs are given in full. The core contribution remains architectural and empirical.
  \item[] Guidelines:
  \begin{itemize}
    \item The answer \answerNA{} means that the paper does not include theoretical results.
    \item All the theorems, formulas, and proofs in the paper should be numbered and cross-referenced.
    \item All assumptions should be clearly stated or referenced in the statement of any theorems.
    \item The proofs can either appear in the main paper or the supplemental material.
    \item Any informal proof provided in the core of the paper should be complemented by formal proofs in appendix or supplemental material.
    \item Theorems and Lemmas that the proof relies upon should be properly referenced.
  \end{itemize}

\item {\bf Experimental result reproducibility}
  \item[] Question: Does the paper fully disclose all the information needed to reproduce the main experimental results of the paper to the extent that it affects the main claims and/or conclusions of the paper (regardless of whether the code and data are provided or not)?
  \item[] Answer: \answerYes{}
  \item[] Justification: The manuscript specifies the architecture, explicit local-consequence construction and preprocessing, training protocol (optimizer, learning rate, batch size, rollout width), credit-assignment rule, checkpoint usage, benchmark aggregation rule, and programmatic CVRPTW data generator. The public code repository provides implementation details, hyperparameter configurations, and reproduction scripts.
  \item[] Guidelines:
  \begin{itemize}
    \item The answer \answerNA{} means that the paper does not include experiments.
    \item If the paper includes experiments, a \answerNo{} answer to this question will not be perceived well by the reviewers.
    \item If the contribution is a dataset and/or model, the authors should describe the steps taken to make their results reproducible or verifiable.
    \item Reproducibility can be provided via code and data, detailed instructions, a hosted model, released checkpoints, or other means appropriate to the research.
    \item While NeurIPS does not require releasing code, submissions must provide a reasonable avenue for reproducibility.
  \end{itemize}

\item {\bf Open access to data and code}
  \item[] Question: Does the paper provide open access to the data and code, with sufficient instructions to faithfully reproduce the main experimental results, as described in supplemental material?
  \item[] Answer: \answerYes{}
  \item[] Justification: A public code repository is provided at \url{https://github.com/Elaina10172004/LINC}, containing LINC training and evaluation code, the programmatic CVRPTW data generator, environment setup, dependency information, evaluation data, selected checkpoints, and reproduction commands.
  \item[] Guidelines:
  \begin{itemize}
    \item The answer \answerNA{} means that paper does not include experiments requiring code.
    \item Please see the NeurIPS code and data submission guidelines for more details.
    \item While release of code and data is encouraged, \answerNo{} is acceptable with a proper justification unless release is central to the contribution.
    \item The instructions should contain the exact command and environment needed to reproduce the results.
    \item The authors should provide instructions on data access and preparation.
    \item The authors should provide scripts to reproduce the main experimental results and baselines when possible.
    \item At submission time, preserve anonymity by releasing anonymized versions if applicable.
  \end{itemize}

\item {\bf Experimental setting/details}
  \item[] Question: Does the paper specify all the training and test details (e.g., data splits, hyperparameters, how they were chosen, type of optimizer) necessary to understand the results?
  \item[] Answer: \answerYes{}
  \item[] Justification: Section~4.1 describes benchmark protocols, training/evaluation sizes, initialization, checkpoint selection, and hardware; Appendix~F gives optimizer, learning rate, batch size, rollout width, and the TSP/CVRP score-morphing schedule.

\item {\bf Experiment statistical significance}
  \item[] Question: Does the paper report error bars suitably and correctly defined or other appropriate information about the statistical significance of the experiments?
  \item[] Answer: \answerYes{}
  \item[] Justification: Main result tables report average distance and gap over fixed multi-instance test sets, with per-instance breakdowns on TSPLIB and Solomon. For the small residual margins on saturated random TSP/CVRP benchmarks, Appendix~E reports paired bootstrap 95\% confidence intervals over test instances for three pre-specified seeds. These intervals capture instance-level variability for fixed trained checkpoints and evaluation protocols; they do not capture the full variance of the entire training process.
  \item[] Guidelines:
  \begin{itemize}
    \item The answer \answerNA{} means that the paper does not include experiments.
    \item Authors should answer \answerYes{} if results are accompanied by error bars, confidence intervals, or statistical significance tests, at least for experiments supporting the main claims.
    \item The factors of variability captured by error bars should be clearly stated.
    \item The method for calculating the error bars should be explained.
    \item It should be clear whether an error bar is a standard deviation or a standard error of the mean.
    \item If error bars are reported in tables or plots, the text should explain how they were calculated.
  \end{itemize}

\item {\bf Experiments compute resources}
  \item[] Question: For each experiment, does the paper provide sufficient information on the computer resources (type of compute workers, memory, time of execution) needed to reproduce the experiments?
  \item[] Answer: \answerYes{}
  \item[] Justification: Training is performed on a single NVIDIA RTX 5090 32GB GPU, with a single training run taking approximately one week. Evaluation is performed on a single NVIDIA A100 40GB GPU, with per-method inference runtimes listed in the result tables. Total compute (including ablation experiments, reduced-budget checks, and repeated-seed evaluations) is approximately 4--6$\times$ the compute of a single training run.
  \item[] Guidelines:
  \begin{itemize}
    \item The answer \answerNA{} means that the paper does not include experiments.
    \item The paper should indicate the type of compute workers, CPU or GPU, internal cluster, or cloud provider, including relevant memory and storage.
    \item The paper should provide compute required for individual experimental runs and estimate total compute.
    \item The paper should disclose whether the full research project required more compute than the reported experiments.
  \end{itemize}

\item {\bf Code of ethics}
  \item[] Question: Does the research conducted in the paper conform, in every respect, with the NeurIPS Code of Ethics?
  \item[] Answer: \answerYes{}
  \item[] Justification: This work studies neural combinatorial optimization algorithms and benchmarks; it does not involve human subjects, sensitive data, or privacy information. The research focuses on improving solution methods for mathematical optimization problems, involving no discrimination, bias, or safety risks. The authors have reviewed the NeurIPS Code of Ethics and confirm full compliance.
  \item[] Guidelines:
  \begin{itemize}
    \item The answer \answerNA{} means that the authors have not reviewed the NeurIPS Code of Ethics.
    \item If the authors answer \answerNo, they should explain any special circumstances requiring deviation from the Code of Ethics.
    \item The authors should preserve anonymity in the explanation where applicable.
  \end{itemize}

\item {\bf Broader impacts}
  \item[] Question: Does the paper discuss both potential positive societal impacts and negative societal impacts of the work performed?
  \item[] Answer: \answerNA{}
  \item[] Justification: This work proposes a decoder architecture for routing optimization and evaluates it on standard benchmarks; it does not raise societal impacts beyond those of existing optimization methods.
  \item[] Guidelines:
  \begin{itemize}
    \item The answer \answerNA{} means that there is no societal impact of the work performed.
    \item If the authors answer \answerNA{} or \answerNo{}, they should explain why the work has no societal impact or why the paper does not address societal impact.
    \item Examples of negative societal impacts include malicious or unintended uses, fairness considerations, privacy considerations, and security considerations.
    \item Authors should consider possible harms from intended use, incorrect results, and misuse.
    \item If there are negative societal impacts, authors may discuss possible mitigation strategies.
  \end{itemize}

\item {\bf Safeguards}
  \item[] Question: Does the paper describe safeguards that have been put in place for responsible release of data or models that have a high risk for misuse (e.g., pre-trained language models, image generators, or scraped datasets)?
  \item[] Answer: \answerNA{}
  \item[] Justification: The paper does not release high-risk pre-trained generative models, scraped web datasets, or other assets requiring special misuse safeguards.
  \item[] Guidelines:
  \begin{itemize}
    \item The answer \answerNA{} means that the paper poses no such risks.
    \item Released models that have a high risk for misuse or dual-use should include necessary safeguards.
    \item Datasets scraped from the Internet could pose safety risks.
    \item Many papers do not require this, but authors should take it into account in good faith.
  \end{itemize}

\item {\bf Licenses for existing assets}
  \item[] Question: Are the creators or original owners of assets (e.g., code, data, models), used in the paper, properly credited and are the license and terms of use explicitly mentioned and properly respected?
  \item[] Answer: \answerYes{}
  \item[] Justification: Major referenced methods, solvers, and benchmark datasets are cited where applicable; the released code documents dataset access, generation procedures, and associated licenses. POMO and SGBS are MIT-licensed. The upstream PolyNet repository reports its GitHub license metadata as ``Other''/``NOASSERTION''; its LICENSE file contains MIT-style permission terms with retained POMO and PolyNet modification notices. The released repository includes third-party notices with URLs, copyright notices, and terms for third-party code, checkpoints, and benchmark data.
  \item[] Guidelines:
  \begin{itemize}
    \item The answer \answerNA{} means that the paper does not use existing assets.
    \item The authors should cite the original paper that produced the code package or dataset.
    \item The authors should state which version of the asset is used and, if possible, include a URL.
    \item The name of the license should be included for each asset where applicable.
    \item If assets are released, license, copyright information, and terms of use should be provided.
    \item For existing datasets that are re-packaged, both the original and derived licenses should be provided.
  \end{itemize}

\item {\bf New assets}
  \item[] Question: Are new assets introduced in the paper well documented and is the documentation provided alongside the assets?
  \item[] Answer: \answerYes{}
  \item[] Justification: The public LINC repository documents the released training/evaluation code, programmatic CVRPTW data generator, selected checkpoints, evaluation data, benchmark conversion files, and reproduction commands. The LINC code is released under the MIT License.
  \item[] Guidelines:
  \begin{itemize}
    \item The answer \answerNA{} means that the paper does not release new assets.
    \item Researchers should communicate details of the dataset/code/model as part of their submissions via structured templates.
    \item The paper should discuss whether and how consent was obtained from people whose asset is used.
    \item At submission time, remember to anonymize assets if applicable.
  \end{itemize}

\item {\bf Crowdsourcing and research with human subjects}
  \item[] Question: For crowdsourcing experiments and research with human subjects, does the paper include the full text of instructions given to participants and screenshots, if applicable, as well as details about compensation (if any)?
  \item[] Answer: \answerNA{}
  \item[] Justification: This work does not involve crowdsourcing experiments or human-subject studies.
  \item[] Guidelines:
  \begin{itemize}
    \item The answer \answerNA{} means that the paper does not involve crowdsourcing nor research with human subjects.
    \item Including this information in supplemental material is fine, but if the main contribution involves human subjects, as much detail as possible should be included in the main paper.
    \item Workers involved in data collection, curation, or other labor should be paid at least the minimum wage in the relevant country.
  \end{itemize}

\item {\bf Institutional review board (IRB) approvals or equivalent for research with human subjects}
  \item[] Question: Does the paper describe potential risks incurred by study participants, whether such risks were disclosed to the subjects, and whether Institutional Review Board (IRB) approvals (or an equivalent approval/review based on the requirements of your country or institution) were obtained?
  \item[] Answer: \answerNA{}
  \item[] Justification: This work does not involve human-subject research and therefore does not require IRB or equivalent approval.
  \item[] Guidelines:
  \begin{itemize}
    \item The answer \answerNA{} means that the paper does not involve crowdsourcing nor research with human subjects.
    \item Depending on the country, IRB approval or equivalent may be required for human-subject research.
    \item For initial submissions, do not include information that would break anonymity, such as the institution conducting the review.
  \end{itemize}

\item {\bf Declaration of LLM usage}
  \item[] Question: Does the paper describe the usage of LLMs if it is an important, original, or non-standard component of the core methods in this research? Note that if the LLM is used only for writing, editing, or formatting purposes and does not impact the core methodology, scientific rigor, or originality of the research, declaration is not required.
  \item[] Answer: \answerNA{}
  \item[] Justification: The core method development and experiments described in this paper do not use LLMs as an important, original, or non-standard research component.
  \item[] Guidelines:
  \begin{itemize}
    \item The answer \answerNA{} means that the core method development in this research does not involve LLMs as any important, original, or non-standard components.
    \item Please refer to the LLM policy in the NeurIPS handbook for what should or should not be described.
  \end{itemize}

\end{enumerate}

\end{document}